\documentclass[lettersize,journal]{IEEEtran}
\usepackage{amsmath,amsfonts}
\usepackage{algorithmic}
\usepackage{algorithm}
\usepackage{array}
\usepackage[caption=false,font=normalsize,labelfont=sf,textfont=sf]{subfig}
\usepackage{textcomp}
\usepackage{stfloats}
\usepackage{url}
\usepackage{verbatim}
\usepackage{graphicx}
\usepackage{cite}
\usepackage{multirow}
\usepackage{array}
\usepackage{caption}
\usepackage{booktabs}
\usepackage{adjustbox}
\usepackage{makecell} 
\usepackage{pifont}
\usepackage{xcolor}

\newcommand{\jc}[1]{\textcolor{black}{#1}}
\long\def\rev#1{{\color{black}#1}}

\begin{document}

\title{Deep Reprogramming Distillation for Medical Foundation Models}

\author{Siyuan Du, Yuhang Zhou, Haolin Li, Jiangchao Yao, Haishuai Wang, Hui Lin, Ya Zhang, Yanfeng Wang
\thanks{S. Du, H. Li are with School of Computer Science, Fudan University and with the Shanghai AI Laboratory, Shanghai, China}
\thanks{H. Wang is with College of Computer Science and Technology, Zhejiang University}
\thanks{Hui Lin is with Sir Run Run Shaw Hospital, School of Medicine, Zhejiang University}
\thanks{Ya Zhang and Y. Wang are with School of Artificial Intelligence, Shanghai Jiao Tong University and with the Shanghai AI Laboratory, Shanghai, China.}
\thanks{Y. Zhou, J. Yao are with the Cooperative Medianet Innovation Center, Shanghai Jiao Tong University and with the Shanghai AI Laboratory, Shanghai, China. (Corresponding authors: Jiangchao Yao. Corresponding e-mails: Sunarker@sjtu.edu.cn)}}

\markboth{Journal of \LaTeX\ Class Files,~Vol.~14, No.~8, August~2021}%
{Shell \MakeLowercase{\textit{et al.}}: A Sample Article Using IEEEtran.cls for IEEE Journals}



\maketitle

\begin{abstract}
Medical foundation models pre-trained on large-scale datasets have shown powerful versatile performance. However, when adapting medical foundation models for specific medical scenarios, it remains the inevitable challenge due to the gap induced by the discrepancy between pre-training and downstream tasks, the real-world computation, and speed constraints. Relevant techniques that probably handle this challenge more or less suffer from some intrinsic limitations. For example, knowledge distillation (KD) assumes that teacher and student models share the same task, training strategy, and model structure family, while prevalent parameter-efficient fine-tuning (PEFT) fails to achieve personalized and lightweight deployment. Even the combination of PEFT and KD still struggles to resolve model structures and training strategies inconsistencies between teacher and student models, leading to inefficient knowledge transfer. In this study, we propose a novel framework called Deep Reprogramming Distillation (DRD) to combat the general adaptation challenge. Specifically, DRD introduces the novel reprogramming module that on the one side overcomes the domain and task discrepancy between pretraining and downstream scenarios, and on the other side builds the student-friendly efficient distillation from foundation models to lightweight downstream models. Furthermore, to mitigate variability under different training conditions, we design a centered kernel alignment (CKA) distillation method to promote robust knowledge transfer. Empirical results show that DRD surpasses previous PEFT and KD methods across 18 medical downstream tasks under different foundation models, covering various scenarios including 2D/3D classification and 2D/3D segmentation.
\end{abstract}


\section{Introduction}
\label{sec:introduction}
%Need modification
\IEEEPARstart{L}{arge-scale} pre-training has promoted the rapid development of medical foundation models, which provides generalizable feature base for various downstream tasks~\cite{lin2023pmc,mei2022radimagenet,nguyen2023lvm,yao2022edge,zhou2024exploring}. 
\jc{Nevertheless, given the downstream-agnostic proxy training strategies (\textit{e.g.,} self-supervised or weakly-supervised learning~\cite{he2022masked,nguyen2023lvm,lin2023pmc}), we usually cannot directly apply the medical foundation models to the specific downstream tasks like tumor benign-malignant diagnosis and rare lung disease diagnosis.}
In addition, for the real-world deployment, \jc{we inevitably need to consider} many constraints regarding model computational limits and inference speed~\cite{zhou2024low}. Therefore, \jc{a downstream adaptation process must be conducted} to mitigate the potential task or domain inconsistencies while ensuring efficient deployment.
% For example, on some edge devices or handheld instruments from different manufacturers, it is not possible to deploy large-capacity foundation models directly.
%不要在这里讲思路。第一段就是为了铺垫研究的background及方向，第二段总结该方向的研究方法，并陈述局限性。第三段讲局限性的粗略设想（这个时候可以引出figure 1的思路）。第四段讲方法设计和why你的方法能够解决问题。我看下面分段不够，请分四段
% As shown in the Figure~\ref{fig:intro1}, foundation models, pretrained on diverse tasks, need to have expertise in handling downstream tasks while reducing deployment costs to enable deployment on resource-limited devices. The general knowledge embedded in foundation models can be leveraged to guide task-specific adaptations during downstream model training. The resulting downstream models retain the necessary task-specific knowledge and their lightweight design allows for broader application in certain medical scenarios. 
% For example, they can be deployed on mobile devices or handheld instruments from various manufacturers, where it is not feasible to directly implement large-capacity foundation models.
% 此处可以选择单栏展示或者多栏展示
\begin{figure*}
    \centering
    \includegraphics[width=0.92\linewidth]{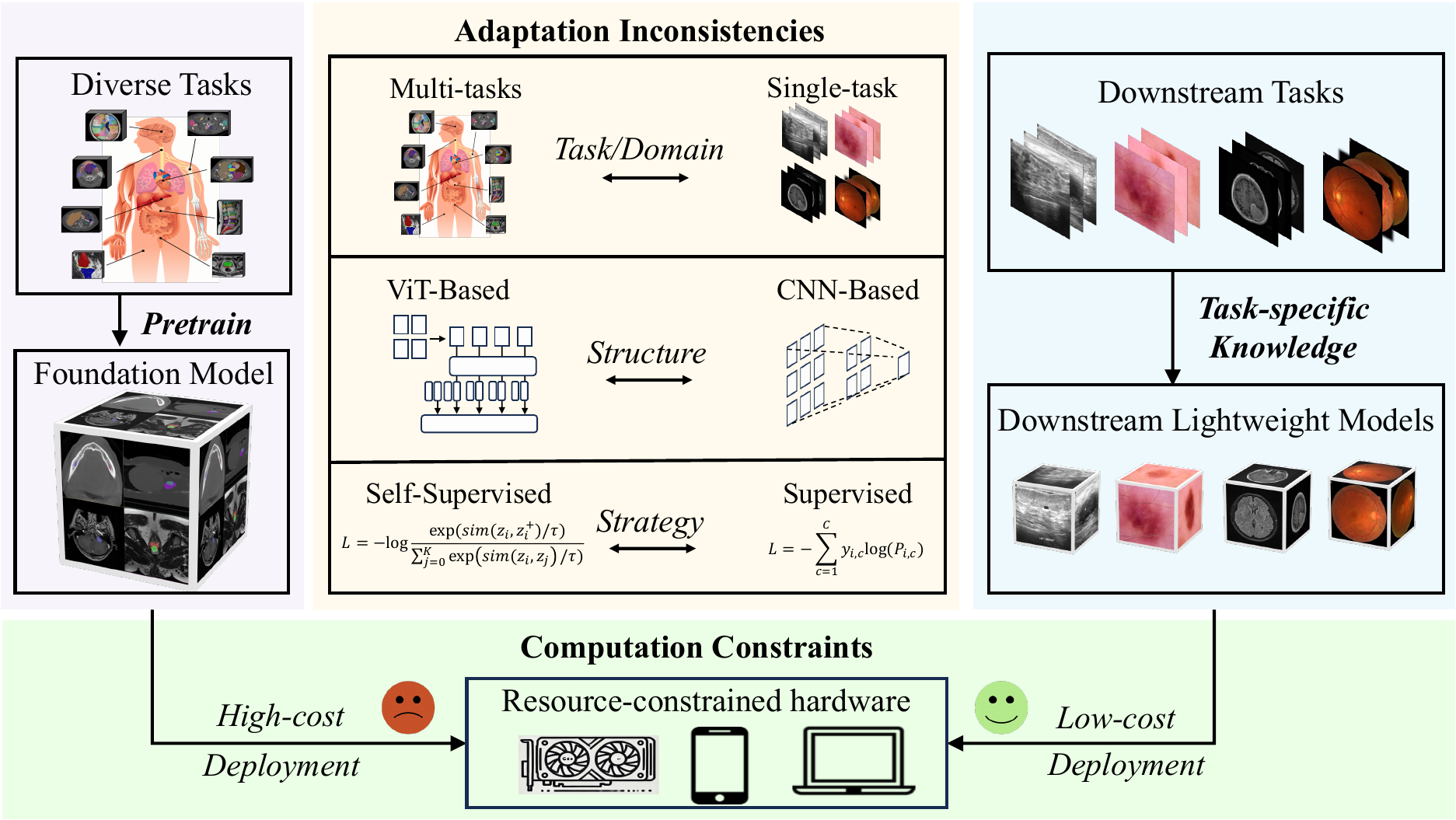}
    \caption{Real-world consideration for the adaptation of medical foundation models. It requires to overcome different inconsistency issues and enables low cost to deploy on resource-constrained devices. The resulting downstream models should well learn task-specific knowledge and build with a lightweight structure to allow for broader application in certain medical scenarios.}
    \label{fig:intro1}
    %\vspace{-0.5cm}
\end{figure*}

% \begin{figure}
%     \centering
%     \includegraphics[width=\linewidth]{img/Introv7.png}
%     \caption{New ways for the adaptation of foundation models. Foundation models require specific knowledge of downstream tasks and lower deployment cost to adapt on resource-constrained devices. }
%     \label{fig:intro1}
%     %\vspace{-0.5cm}
% \end{figure}

Currently, the mainstream downstream adaptation methods for foundation models are parameter-efficient fine-tuning (PEFT)~\cite{xin2024parameter, chen2022adaptformer, jia2022visual, li2021prefix, hu2021lora,he2023parameter}, which leverage a small number of trainable parameters to achieve performance comparable to full fine-tuning. 
Despite the reduction in training cost, these methods might not meet the deployment constraints for memory usage and inference speed in practice.
Knowledge distillation (KD)~\cite{hinton2015distilling,tung2019similarity,tian2019contrastive,park2019relational,yang2022masked,liu2023norm,li2024lorkd}, as an effective means of knowledge transfer, can compress the model size without sacrificing much performance of teacher models. 
However, unlike the \jc{ordinary} knowledge distillation scenario, the pre-training data used by medical foundation models might be task- or domain- inconsistent with the downstream data, leading to a lack of direct correlation between the foundation model and the student model, thereby weakening the adaptation gain of the foundation model. 
A naive integration of KD and PEFT involves pre-tuning the foundation model with PEFT for downstream tasks before applying KD. However, this still disregards the model structures and training strategies inconsistencies between the foundation and downstream models, which hinders the transfer of student-friendly knowledge \jc{and thus limits the improvements.}
%
% The recent exploration through model reprogramming~\cite{xu2023towards} mainly focuses on natural scenarios, while we extend this idea into the medical scenarios with the technical improvement for an end-to-end training.

As shown in Fig.~\ref{fig:intro1}, \jc{when adapting medical foundation models to downstream tasks, we actually need to consider several inconsistency issues while reducing deployment cost on resource-limited devices.}
\jc{Basically, we can transfer} the general knowledge embedded in medical foundation models to guide the training of a lightweight downstream model.
However, the gap between tasks, model structures, and training strategies often results in non-negligible negative impact when transferred naively.
To guarantee the effectiveness of knowledge transfer to the downstream model, it is crucial to refine the potentially biased knowledge in medical foundation models with mediating the inconsistencies related to tasks, structural designs, as well as training strategies \textit{etc}.

% However, in this process, previous methods such as PEFT and KD partially overlook the disparities between the foundation model and the downstream model in terms of tasks, model structure families, and training strategies. 
%如图，这里可以加一个CKA对比图，
%如图，我们可以看到在特征层面，PEFT与KD方法并没有有效对齐foundation model与downstream model的偏差（1.FD/DS; 2.FD+PEFT/DS; 3.FD+KD/DS; 4.FD+DRD/DS）
% Consequently, the general knowledge transferred is inadequately adapted to the downstream model, leading to suboptimal knowledge transfer.
%
Specifically, to address the main challenge, we propose an effective framework called deep reprogramming distillation (DRD) for adapting  medical foundation models on downstream tasks.
Firstly, we introduce a reprogramming module to mitigate inconsistency, which helps extract features more relevant to downstream tasks. Here, we treat the foundation model as a basic stack of blocks with fixed parameters.
% , which avoids direct manipulation of the internal parameters like PEFT and circumvents the significant GPU costs associated with backpropagation in the large foundation model. 
%
Secondly, to ensure that reprogrammed features can be smoothly mimicked by downstream models, we adopt a deep co-training mechanism to establish connections between the reprogrammed knowledge and the knowledge extracted by student models, and encourage them to learn similar feature representation through the shared student branches. 
Additionally, we introduce the Centered Kernel Alignment (CKA) distillation to promote robust knowledge transfer, so as to alleviate the randomness introduced due to the various training conditions. 
%这里需不需要修改一下，像昊林那篇一样？
This paper is an extension of our preliminary work~\cite{zhou2024reprogramming}, which introduced model reprogramming to address task bias during adaptation and utilized CKA distillation to facilitate robust feature transfer.
The contributions of this paper on top of \cite{zhou2024reprogramming} are summarized as follows.
\begin{itemize}
\item  We design a new framework, Deep Reprogramming Distillation, to facilitate downstream adaptation of medical foundation models.
%
% DRD overcomes inconsistencies of task, model structure and training strategies \textit{etc.}, enabling lightweight deployment and efficient knowledge transfer, which cannot be achieved by previous methods.
\rev{DRD targets a broader practical setting than previous methods by jointly handling task/domain discrepancy, teacher--student structural discrepancy, and lightweight deployment in one stage.}
\item 
We propose a novel component for DRD, Deep Co-training Reprogramming, which aligns tasks, modalities and structures for easier feature mimic by downstream models to promote more robust knowledge transfer.
% , and CKA distillation, which enhances the robust feature transfer across the various training conditions.
\item 
We select eight types of medical foundation models to conduct extensive experiments on eighteen public datasets, encompassing various tasks and modalities, and the results prove that DRD consistently improves the performance across different training conditions. 
\end{itemize}

% Motivation:
% In previous studies, teachers and students with similar structures and identical training strategies naturally exhibit similar learned representations in the feature space. However, when the structures and training strategies of teachers and students are heterogeneous, the features extracted occupy different potential feature spaces, as illustrated in Figure ~\ref{fig:intro1}. Under these conditions, simple feature alignment methods for distillation, such as mean square error (MSE) loss, prove ineffective and can even negatively impact the learning of students' networks. (need a table to support it.)
% PEFT与KD方法无法提取出下游对下游有效的知识

\section{Related Work}
% \subsection{Medical Foundation Models}
% Medical foundation models are a class of pre-trained models tailored for biomedical imagery and data, designed to enhance the efficiency and accuracy of subsequent tasks. These models leverage extensive pre-training on large-scale medical datasets to capture complex biomedical signals and patterns, supporting a variety of clinical applications. For instance, some classification foudation Models ~\cite{lin2023pmc,mei2022radimagenet,nguyen2023lvm} are trained on large-scale datasets through supervised, self-supervised, or contrastive learning methods. MedSAM~\cite{ma2024segment} finetuned the Segment Anything Model(SAM)~\cite{kirillov2023segment} on medical datasets to obtain a foundation model for medical segmentation with superior performance and generalization capability. SAT 
% (MedSAM,MSA,SAMMED);

\subsection{Knowledge Distillation}

Knowledge Distillation \jc{origins} for model compression, which utilizes a smaller ``student'' model to mimic a larger ``teacher'' model. 
This concept was initially introduced by Bucila et al.~\cite{bucilua2006model} and later enhanced by Hinton et al.~\cite{hinton2015distilling}. Subsequent research has advanced logits-based Knowledge Distillation through model ensembles~\cite{malinin2019ensemble}, contrastive learning techniques~\cite{tian2019contrastive} or by means of incorporation of structural information~\cite{park2019relational}.
Other stream of approaches utilize intermediate features as guiding knowledge. For example, Romero et al.~\cite{romero2014fitnets} introduced FitNets, which minimised the discrepancy between the projected feature and the teacher feature using mean square loss. 
Zagoruyko et al.~\cite{zagoruyko2016paying} proposed an attention transfer mechanism in which the student learns to mimic the attention maps of the teacher.
Furthermore, Some studies aimed to address how to distil efficiently when student and teacher architectures are heterogeneous. For instance, 
% Park et al.~\cite{park2021learning} adopted a student-aware teacher learning procedure before feature distillation to make knowledge transfer process more efficient. 
Hao et al.~\cite{hao2024one} transferred the mismatched representations into the aligned logits space by incorporating additional exit branches into the student model before matching the outputs with logits from teacher.
%
% Although existing distillation methods have achieved remarkable performance, they assume that students and teachers have the same task or the same model structure family. When tasks, structures or even training strategies are heterogeneous, existing methods may fail in transferring knowledge because of the huge deviation in the features of the teacher and the student.
Although existing distillation methods have achieved remarkable performance, they typically assume that the student and teacher models share the same task or belong to the same model family. However, when tasks, model structures, or even training strategies differ significantly, they will struggle to transfer knowledge effectively due to the large discrepancy between the teacher and student features.

\subsection{Parameter-Efficient Fine-Tuning (PEFT)}
% Parameter-Efficient Fine-Tuning (PEFT) methods allow for the adaptation of pre-trained models to specific tasks by fine-tuning only a small fraction of the model's parameters. Current PEFT techniques are grouped into three main types. Adapter-based methods, the first type, introduce additional trainable components into the original frozen backbone ~\cite{houlsby2019parameter,he2021towards,mahabadi2021parameter,karimi2021compacter}. The second category, Prompt-based methods, incorporate additional soft tokens (prompts) to the initial input and solely fine-tune these trainable parameters~\cite{lester2021power,razdaibiedina2023residual,jia2022visual,dong2022lpt}. The third category includes LoRA~\cite{hu2021lora} and its variants\cite{liu2024dora,qiu2023controlling,yeh2023navigating}, which are  notable for not increasing the inference burden. These techniques employ low-rank matrices to simulate modifications in weights during fine-tuning and can merge with pre-trained weights before inference.Further, the use of PEFT in medical imaging has been expanded to include  methods like MeLo \cite{zaken2021bitfit}, which adopts low-rank adaptation instead of resourcedemanding fine-tuning to enable the development of a single CAD model for multiple clinical tasks in a lightweight manner. Wu et al. ~\cite{wu2023medical}introduce a specialized adapter module integrated into Segmentation Anything Model (SAM) to adapt SAM in medical segmentation tasks.

Parameter-Efficient Fine-Tuning methods facilitate the adaptation of pre-trained models to specific tasks by fine-tuning a small fraction of parameters, which can be roughly categorized into three types. 
First, adapter-based methods involve introducing additional trainable components into the frozen backbone~\cite{houlsby2019parameter,he2021towards,mahabadi2021parameter,karimi2021compacter}. 
Second, prompt-based methods incorporate additional soft tokens (prompts) into the initial input, focusing on fine-tuning these specific parameters~\cite{lester2021power,razdaibiedina2023residual,jia2022visual,dong2022lpt}. 
The third category includes LoRA~\cite{hu2021lora} and its variants~\cite{liu2024dora,qiu2023controlling,yeh2023navigating}, which are particularly notable for not increasing the inference burden.
These techniques utilize low-rank matrices to simulate weight modifications during fine-tuning, allowing them to be seamlessly integrated with pre-trained weights before inference.
Moreover, the application of PEFT in medical imaging has expanded to include methods such as MeLo~\cite{zaken2021bitfit}, which employs low-rank adaptation in medical scenarios, enabling the development of a single CAD model for multiple clinical tasks in a lightweight manner. 
Wu et al.~\cite{wu2023medical} introduced a specialized adapter module into the Segmentation Anything Model (SAM) to adapt it for medical segmentation tasks.
Although PEFT methods reduce the training costs for adaption, it cannot decrease the practical deployment costs, making them unsuitable for resource-limited scenarios.

% Although PEFT methods can reduce the training cost when adapting foundation models to downstream tasks, it does not lower the actual deployment cost, making it unsuitable for use in resource-constrained downstream scenarios.
\section{Deep Reprogramming distillation}
% In this section, we first demonstrate the preliminary required for our methods. Then we present each of the three key technical points included in our approach: Co-training Reprogramming, Centered Kernel Alignment Distillation, and Deep inverse-teaching Co-training. Finally, we describe the specific network architecture and implementation details of our RD approach and the DRD approach.

In this section, we initially present the preliminary and motivation, followed by a detailed description of three key components of our method: Reprogramming Module, Deep Co-training Reprogramming, and Centered Kernel Alignment Distillation. Finally, we give the loss functions and discuss the advantages of DRD in comparison with previous approaches.

\subsection{Preliminary}
Let $T$ represents the medical foundation model pre-trained on broad data, namely, the teacher model, and $S$ represents the lightweight model to be deployed, namely, the student model. 
In the downstream adaptation, medical foundation models may have different structures, and the downstream tasks may also be segmentation or classification.
%
% For simplicity, we group the layers of both $T$ and $S$ into a fixed number of $N$ blocks (to be aligned in DRD) and retain the output head of $S$. 
\rev{For simplicity, we group the layers of both $T$ and $S$ into a fixed number of $N$ blocks, which define a coarse stage-level correspondence in DRD, and retain the output head of $S$. This correspondence only specifies coarse stage-wise injection points, rather than assuming a priori semantic equivalence between the $i$-th blocks of $T$ and $S$.}
Note that, the layer size of each block in $T$ and $S$ can vary. 
We denote $B^T = \{B_i^T\}_{i=1}^N$ as the teacher's blocks, $B^S = \{B_i^S\}_{i=1}^N$ as the student's blocks, and $H^S$ as the output head of $S$.
Given the downstream data $D = \{x_i, y_i \}_{i=1}^M$, where $M$ is the training data size, $x_i$ represents the input image, and $y_i$ represents the label in classification tasks or the mask in segmentation tasks. Our goal is to transfer task-relevant knowledge from $T$ to $S$, enabling $S$ to use foundation model knowledge while ensuring memory and computational efficiency during inference.
% Our goal is to transfer the knowledge relevant to downstream tasks from $T$ to $S$, so that $S$ can not only leverage the knowledge in foundation models but also ensure efficiency in memory and computation for inference.

% We propose a new reprogramming distillation framework named RD which focuses on a more general scenario where the structures and data distribution may be different between upstream and downstream.
% In order to mitigate the bias caused by the network structure and training strategy even further. We propose an improvement on RD by proposing its method DRD, which reshapes the teacher's output features to be more student-frendly by adding student network branching assistance in the reprogramming process.

% We propose a new reprogramming distillation framework, termed DRD, which addresses a more general scenario where structures and data distributions differ between upstream and downstream tasks. To further mitigate bias introduced by network structure and training strategy, we introduce an enhancement to RD—dubbed DRD. This method reshapes the teacher's output features to be more student-friendly by incorporating student network branching assistance during the reprogramming process.

\subsection{Motivation}\label{sec:moti}
The size of medical foundation models has steadily increased over time~\cite{azad2023foundational}, significantly limiting their deployment in downstream medical scenarios due to computational resource constraints.
Simultaneously, the tasks and domains covered by the training datasets of these models are becoming more expansive~\cite{zhang2024challenges}. While this enhances their versatility and generalization, the broad scope of knowledge they encompass may lack the necessary expertise for downstream medical tasks. Consequently, their performance sometimes is inferior to that of lightweight models independently trained on downstream data. 
Through a series of preliminary experiments, we observed that large-scale foundation models like MedSAM fall short in specialized downstream medical tasks. Specifically, as shown in the Table~\ref{tab:motivation1}, despite the large scale and the richness of pre-training tasks (84 segmentation tasks across 10 modalities), MedSAM underperforms on downstream tasks compared to several lightweight models.

\begin{table}[t!]
  \centering
  \caption{Performance comparison between foundation model MedSAM and lightweight models on downstream task \jc{datasets} FIVES~\cite{jin2022fives}, USC~\cite{USC2013}, CDPRD~\cite{humans_in_the_loop_2023}.}
  \setlength{\tabcolsep}{3pt}{
    \begin{tabular}{c|c|c|ccc}
    \toprule[1.2pt]
    Model & Params (M) & Training Tasks & FIVES & USC   & CDPRD \\
    \midrule
    MedSAM & 93.6  & 84 diverse tasks & 16.53  & 83.73  & 78.34  \\
    \midrule
    ViT-Tiny & 6.2   & \multirow{4}[2]{*}{Downstream task}   & 83.06  & 87.93  & 90.91  \\
    ResNet18 & 15.9  &  & 68.00  & 85.32  & 87.61  \\
    ShuffleNet & 6.1   &       & 56.25  & 86.19  & 76.54  \\
    MobileNet & 7.2   &       & 69.58  & 86.47  & 84.04  \\
    \bottomrule[1.2pt]
    \end{tabular}}%
  \label{tab:motivation1}%
\end{table}%

\begin{table}[t!]
  \centering
  \caption{ Performance changes of downstream student models using different adaptation methods, with or without model structural inconsistencies, compared to vanilla-trained models \jc{on FIVES dataset}.}
  \setlength{\tabcolsep}{3pt}{
    \begin{tabular}{c|c|c|ccc}
    \toprule[1.2pt]
    Teacher & Student & Structure Change & Hint  & MobSAM & Lora+KD \\
    \midrule
    \multirow{4}[4]{*}{MedSAM} & ViT-Tiny & {$\texttt{\textbf{ViT}}\rightarrow\texttt{\textbf{ViT}}$} & 2.2$\uparrow$ & 2.2$\uparrow$ & 3.8$\uparrow$ \\
\cmidrule{2-6}          & ResNet18 & \multirow{3}[2]{*}{$\texttt{\textbf{ViT}}\rightarrow\texttt{\textbf{CNN}}$} & 4.1$\downarrow$ & 8.4$\downarrow$ & 5.0$\uparrow$ \\
          & ShuffleNet &       & 0.1$\uparrow$ & 7.8$\downarrow$ & 11.0$\uparrow$ \\
          & MobileNet &       & 4.3$\downarrow$ & 11.8$\downarrow$ & 1.5$\downarrow$ \\
    \bottomrule[1.2pt]
    \end{tabular}}%
  \label{tab:motivation2}%
\end{table}%

% \begin{table}[t!]
%   \centering
%   \caption{Performance changes of downstream student models using different adaptation methods, with or without model structural inconsistencies,  compared to vanilla-trained models (+: performance improvement; -: performance decrease). The teacher model is MedSAM (ViT-based), and the student models are: ViT-Tiny (ViT-based), ResNet18, ShuffleNet, and MobileNet (CNN-based). The dataset used is FIVES.}
%     \setlength{\tabcolsep}{3pt}{
%     \begin{tabular}{c|c|c|ccc}
%     \toprule[1.2pt]
%     Teacher & Student & Model Change & Hint  & MobSAM & Lora+KD \\
%     \midrule
%     \multirow{4}[4]{*}{MedSAM} & ViT-Tiny & {$\texttt{\textbf{ViT}}\rightarrow\texttt{\textbf{ViT}}$} & +2.2 & +2.2 & +3.8 \\
% \cmidrule{2-6}          & ResNet18 & \multirow{3}[2]{*}{$\texttt{\textbf{ViT}}\rightarrow\texttt{\textbf{CNN}}$} & -4.1 & -8.4 & +5.0 \\
%           & ShuffleNet &       & +0.1 & -7.8 & +11.0 \\
%           & MobileNet &       & -4.3 & -11.8 & -1.5 \\
%     \bottomrule[1.2pt]
%     \end{tabular}}%
%   \label{tab:motivation2}%
% \end{table}%

Basically, existing KD and PEFT approaches might be used for adaptation. 
However, as discussed in the introduction section, these methods derived from KD and PEFT also suffer from some limitations. Most KD methods assume that both the teacher (foundation model) and the student (downstream model) share the same task, employ similar training strategies, and belong to the same model family, which limits the performance in scenarios where the downstream model deviates from the foundation model's task or architecture to meet specific clinical requirements.
PEFT methods typically fall short in enabling the lightweight model deployment across diverse real-world medical scenarios. 
Even attempts to integrate PEFT with KD have been inadequate in addressing model structures and training strategies inconsistencies between teacher and student models, resulting in inefficient knowledge transfer.
As shown in the Table~\ref{tab:motivation2}, KD methods like Hint and MobSAM, PEFT methods like LoRA combined with KD, can enhance student model performance when both the teacher and student belong to the same model family. However, when the teacher and student models are heterogeneous, these methods often lead to the negative outcomes.
\rev{These results suggest that directly applying existing KD/PEFT paradigms is insufficient for heterogeneous foundation-model adaptation, where the transferred knowledge should be aligned with the downstream task and student architecture.}

\begin{figure*}[t!]
    \centering
    \includegraphics[width=0.92\linewidth]{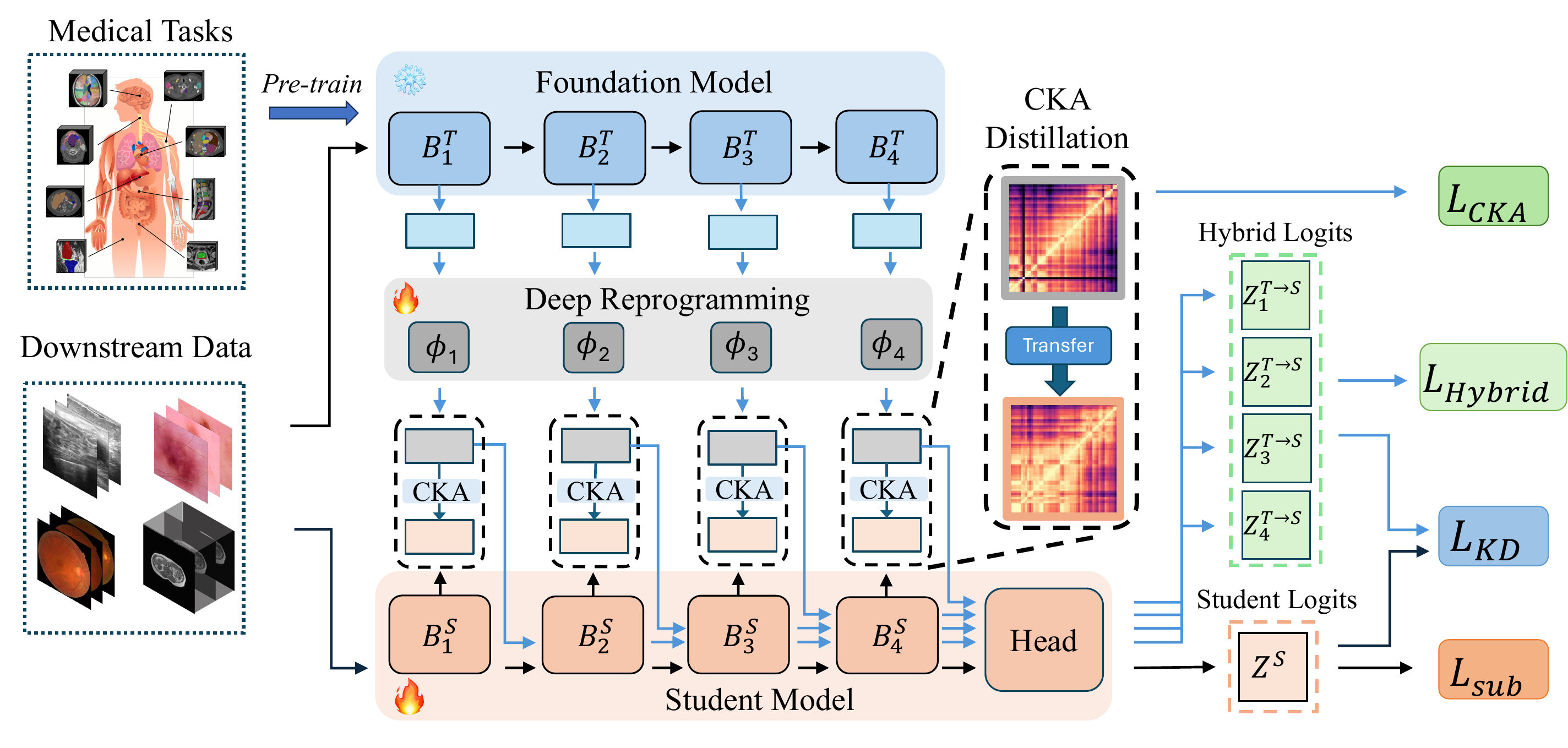}
    \caption{The overall framework of Deep Reprogramming Distillation. During training, the foundation model is frozen. The deep reprogramming module reprograms the intermediate features of the foundation model through the student model branches, with downstream data supervising the co-training process. Additionally, the student model further benefits from CKA distillation for enhanced feature alignment and hybrid student logits distillation for effective logits-level knowledge transfer.}
    \label{fig:method}
    %\vspace{-0.5cm}
\end{figure*}

\jc{The drawbacks of existing methods \textit{w.r.t.} the inconsistency issues during adaptation motivate us to design a new framework.}
Specifically, to mitigate inconsistencies related to task, model structures, and training strategies, we build reprogramming blocks to establish connections between the teacher
and the student model. 
%
% We implement a novel co-training reprogramming that encourages similar and task-relevant features to be learned through shared branches during adaptation. 
\rev{We implement a novel co-training reprogramming strategy, where the reprogramming blocks are jointly optimized with downstream supervised learning, enabling teacher features to be reshaped into task-relevant and student-compatible representations during adaptation.}
Furthermore, to achieve a lightweight downstream student model, we incorporate both centered kernel alignment distillation and multi-level logits distillation, which efficiently transfer the reprogrammed knowledge to the student model.
Our DRD framework is illustrated in Fig.~\ref{fig:method}, and in the following, we discuss the key components concretely.

% %\vspace{-0.78cm}
\subsection{Reprogramming Module}
To mitigate aforementioned challenge, we introduce the reprogramming module to re-purpose the feature space of the teacher model, which is inspired by the recent model reprogramming study~\cite{chen2022model,xu2023towards}. By adding proper transformation layers, it promotes the cross-domain adaptation. Despite effectiveness~\cite{chen2022model}, our goal here extends both domain and task prototypes. We aim to enable the lightweight deployment of adapted models to specific medical tasks, which requires to extract task-relevant knowledge for efficient transfer.

For clarity, the feature maps that are output from $B^T$ and $B^S$ are represented $\{ f_i^T \}_{i=1}^{N},\:\enspace f_i^T \in \mathbb{R}^{C \times H \times W}$ and $\{ f_i^S \}_{i=1}^{N},\:\enspace f_i^S \in \mathbb{R}^{C' \times H' \times W'}$, where $C, C'$ denote channel numbers, and $H, W$, $H', W'$ are the dimensions of the feature maps.
We have designed a set of feature reprogramming blocks along with corresponding teacher blocks, denoted as $\{\phi_i(\cdot)\}_{i=1}^{N}$, to perform the mapping and resizing of the teacher's features, with the objective of reprogramming the teacher’s knowledge at the feature level. 
These blocks employ a trainable convolutional network that adjusts the channels and spatial dimensions of the teacher features $\{ f_i^T \}_{i=1}^{N}$ to align with the student features $\{ f_i^S \}_{i=1}^{N}$. 
In implementation, each $\phi_i(\cdot)$ is instantiated as a lightweight three-layer convolutional projector.
For the $i$-th block, the reprogrammed feature is defined as
\[ f_i^{T \rightarrow S} = \phi_i(f_i^T), \] 
where $\phi_i$ is the reprogramming function that transforms the teacher's feature map $f_i^T$ to match the student's feature map $f_i^S$ in dimensions and assist the training of the student model.

\subsection{Deep Co-training Reprogramming}
Considering that different model structures and training strategies may cause different tendencies in feature extraction, we propose a novel co-training mechanism to ensure that the reprogrammed features can be smoothly mimicked by the student model. In this mechanism, the output features from each block of the teacher network are reprogrammed and fed into the corresponding blocks of the student network. 
% instead of the subsequent blocks of the teacher network.
%
% For example, if both teacher and student networks are divided into four blocks, the output feature map from the first block of the teacher network, after reprogramming, is directly passed to the second, third, fourth blocks 
%  and output head of the student network to obtain the final prediction.
%

Formally, the downstream image input $x$ is processed by the reprogramming module after being input into $T$. The reprogrammed feature $f_i^{T \rightarrow S}$ from the $i$-th block is fed into the subsequent student blocks $B_{i+1}^S, B_{i+2}^S, \dots, B_N^S, H^S$ to compute the final logits, named hybrid logits and denoted as $z_i^{T \rightarrow S}$. Meanwhile, the input $x$ is also fed into $S$ to obtain the student final logits $z^S$. Subsequently, we compute the loss of the two types of logits: $\{z_i^{T \rightarrow S}\}_{i=1}^N$ and $z^S$, using the same loss function to measure the discrepancy with the ground truth $y$. 
Additionally, we use the Kullback-Leibler divergence to align $\{z_i^{T \rightarrow S}\}_{i=1}^N$ and $z^S$, in order to multi-level distill the student-friendly knowledge, reprogrammed from the teacher, into the student at the logits level. This co-training mechanism utilizes downstream data and student models to facilitate the feature reprogramming process of the teacher model. It ensures that the knowledge transferred to the student models encompasses both the robust feature extraction capabilities inherent in the teacher model and the adaptability to the downstream data and the student model architecture. 
\rev{Notably, rather than hand-crafted fine-grained matching, DRD relies only on a coarse stage decomposition, while the final cross-model alignment is progressively learned by the reprogramming blocks through end-to-end training.}

% This approach thus promoting effective knowledge transfer and enhancing the student performance on downstream tasks.

\subsection{Centered Kernel Alignment Distillation}
In the previous subsection, we can obtain the reprogrammed student-friendly teacher features and perform the teacher-to-student knowledge transfer at the logit level. \jc{Here, we build the knowledge distillation at the feature level to improve the efficiency.} Specifically, during downstream adaptation, various training conditions such as different model structures, random seeds, etc., may introduce noise or uncertainty in feature distillation~\cite{ge2024discrepancy}. For robustness, we explore using Centered Kernel Alignment (CKA) distillation \textit{w.r.t.} the correlation and higher-order information in features to reduce training instability~\cite{kornblith2019similarity}.

As aforementioned, $\{f_i^{T \rightarrow S}\}_{i=1}^N$ denotes the extracted features after reprogramming blocks, $f_i^S$ denotes features from student model. 
We use $K_i = (f_i^{T \rightarrow S}) (f_i^{T \rightarrow S})^\top$ and $L_i = (f_i^S) (f_i^S)^\top$ to represent the pair-wise feature similarities. 
Let $\boldsymbol{H}=\boldsymbol{I}_n-\frac1n\boldsymbol{1}\boldsymbol{1}^\mathsf{T}$ be the centering matrix and $K^{\prime}_i= HK_iH$, $L^{\prime}_i = HL_iH$, where $n$ is the batch size. Then, the similarity of the centered similarity matrices can be measured by $\operatorname{HSIC}({K_i},{L_i})=\frac{{K_i}^{\prime}\cdot{L_i}^{\prime}}{(n-1)^2}$~\cite{kornblith2019similarity}.
Note that, although $\operatorname{HSIC}$ is invariant to orthogonal transformations of features, it is not invariant to isotropic scaling~\cite{kornblith2019similarity}. To address this, we normalize HSIC and then construct the CKA loss as follows,
\begin{equation}
\mathcal{L}^{CKA}_i=-\frac{\mathrm{HSIC}({K_i},{L_i})}{\sqrt{\mathrm{HSIC}({K_i},{K_i})\cdot\mathrm{HSIC}({L_i},{L_i})}}, 
\end{equation}
which can be used to determine the correspondence between hidden layers of networks with different random initialization and widths~\cite{nguyen2020wide}.
% The training loss of DRD can be written as
% \begin{equation}
% \mathcal{L}_{train} = \mathcal{L}_{CE} + \alpha \mathcal{L}_{hybrid} + \beta \mathcal{L}_{KL}+\mathcal{L}_{CKA}(f_t, f_s)). 
% \end{equation}
Here, CKA feature alignment can be utilized not only as a feature distillation method to transfer the teacher's reprogrammed information to the students at the feature level, but also as a constraint for co-training reprogramming to achieve robust similarity between the two.
\rev{Maximizing} feature similarity during co-training reprogramming helps establish a strong starting point for downstream models and adapts features for downstream structure early, resulting in more suitable transfer feature patterns.

\subsection{Training Objective}
The training process is driven by two types of loss functions: 1) One type is the supervised learning loss for the downstream task, which ensures the performance of the student model and directs the reprogramming of the foundation model. For classification tasks, cross-entropy loss is used, while DICE loss~\cite{dice1945measures} is applied for segmentation tasks. 2) The second type is the distillation loss, which is designed to efficiently transfer knowledge from both the logits and feature levels.
\begin{itemize}
    \item \textbf{Student Supervised Learning Loss}: The cross-entropy loss or Dice loss between the student’s final logits $z^S$ and the ground truth $y$, defined as:
    \[
    \mathcal{L}_{sup} = CE(z^S, y) \quad \text{or} \quad Dice(z^S, y)
    \]
    \item \textbf{Hybrid Supervised Learning Loss}: The cross-entropy loss or DICE loss calculated between each of the $N$ hybrid logits $z_i^{T \rightarrow S}$ and the ground truth $y$, defined as:
    \[
    \mathcal{L}_{hybrid} = \sum_{i=1}^N CE(z_i^{T \rightarrow S}, y)\quad \text{or} \quad \sum_{i=1}^N Dice(z_i^{T \rightarrow S}, y)
    \]
    \item \textbf{Hybrid Logits Distillation Loss}: The Kullback-Leibler divergence loss between the hybrid logits and the student’s logits, defined as:
    \[
    \mathcal{L}_{KD} = \sum_{i=1}^N KL(z_i^{T \rightarrow S}, z^S).
    \]
    \item \textbf{CKA Features Distillation Loss}: The CKA loss, ensuring robust feature-level alignment, as defined earlier:
    \[
    \mathcal{L}_{CKA}=\frac{1}{N}\sum_{i=1}^N\mathcal{L}^{CKA}_i.
    \]
\end{itemize}

The total loss function is a weighted sum of these losses:
\[
\mathcal{L}_{train} = \mathcal{L}_{sup} + \alpha \mathcal{L}_{hybrid} + \beta \mathcal{L}_{KD} + \mathcal{L}_{CKA},
\]
where $\alpha$ and $\beta$ are hyperparameters that balance the contributions of each loss component. 

\newcommand{\cmark}{\textcolor{teal}{\ding{51}}} % Check mark
\newcommand{\xmark}{\textcolor{red}{\ding{55}}}  

%这里出现的方法需要加上引用吗？都是第一次出现的方法。
\begin{table}[t!]
  \centering
  \caption{Comparison of different methods for medical foundation model downstream adaptation from different aspects. Here, MeLo* combines MeLo and KD.}
  \setlength{\tabcolsep}{2pt}{
    \begin{tabular}{c|c|c|cc|cc}
    \toprule[1.2pt]
    Methods & Eff. Deploy & One-stage & Cla.  & Seg.  & Task Diff. & Struct. Diff. \\
    \midrule
    Hint\cite{romero2014fitnets}  &\cmark &\cmark &\cmark &\cmark &\xmark &\xmark \\
    \rev{OFA~\cite{hao2024one}} &\cmark &\cmark &\cmark &\xmark &\xmark &\cmark \\
    CIRKD\cite{yang2022cross} &\cmark &\cmark &\xmark &\cmark &\xmark &\xmark \\
    MobSAM\cite{zhang2023faster} &\cmark &\cmark &\xmark &\cmark &\xmark &\cmark \\
    \midrule
    MeLo\cite{zhu2023melo}  &\xmark &\cmark &\cmark &\cmark &\cmark &\xmark \\
    MeLo* &\cmark &\xmark &\cmark &\cmark &\cmark &\xmark \\
    \midrule
    Ours  &\cmark &\cmark &\cmark &\cmark &\cmark &\cmark \\
    \bottomrule[1.2pt]
    \end{tabular}%
  \label{tab:method_compare}}%
\end{table}%

\subsection{Discussion on advantages}
As illustrated in Table~\ref{tab:method_compare}, \rev{DRD addresses a broader practical adaptation setting than previous downstream adaptation methods, covering lightweight deployment, task diversity, and structural discrepancy. Specifically, DRD has several advantages:}
1) The model obtained after downstream adaptation is lightweight, and the structure can be customized according to medical scenarios, making deployment more flexible; 2) Since the trainable parameters are completely isolated from the medical foundation model, there is no need to backpropagate gradients through the backbone of medical foundation model during training, which actually reduces the GPU usage; 3) The framework exhibits the strong generalization, providing downstream adaptation solutions for various foundation models in both classification and segmentation scenarios;
4) Its distinctive co-training framework mitigates the inconsistencies between the teacher and student models, and its CKA promotes an end-to-end training to avoid the heuristic stage-wise training~\cite{xu2023towards}, facilitating more effective knowledge transfer.

\section{Experimental Setup} \label{experiment}
% In this section, we provide the comprehensive experimental verification of our method and in-depth analysis.
In this section, we detail the experimental setup, including datasets, evaluation metrics, baselines, implementation details, and the models used.
% conduct a range of experiments \textit{w.r.t.} downstream classification and segmentation tasks in medical domain. First, we will describe the experimental setups including datasets, evaluation metrics and baselines. Then, we will present the chosen medical foundation models and the corresponding downstream models along with the implementation details for experiments. Finally, we will show the results in each task, followed by the ablation and further analysis.

\subsection{Datasets}
To comprehensively verify DRD, 
we selected eighteen medical image datasets with various modalities and tasks for downstream adaptation.
\textbf{2D classification datasets:} BUSI~\cite{al2020busi}, ISIC~\cite{tschandl2018ISIC}, Covid~\cite{xingyi2020covid_CT}, BTC~\cite{saleh2020BTC} and ChestXray~\cite{cohen2020covid_xray}.
\textbf{3D classification datasets:} 
MosMedData (MMD)~\cite{morozov2020mosmeddata}, Lung Adenocarcinoma Classification (LAC)~\cite{LAC2020}.
%https://www.kaggle.com/datasets/mathurinache/mosmeddata-chest-ct-scans-with-covid19/data 
%
\textbf{2D segmentation datasets:} CoNIC~\cite{graham2024conic,graham2021lizard}, CDPRD~\cite{humans_in_the_loop_2023}, FIVES~\cite{jin2022fives},  UWaterloo Skin Cancer (USC)~\cite{USC2013}, TN3K~\cite{gong2021multi}, 
% Mandibles~\cite{abdi2015automatic},
Robotool (Robo)~\cite{garcia2021image}, 
CHAOS~\cite{kavur2019,CHAOS2021,CHAOSdata2019}. 
The CHAOS dataset is derived by slicing original 3D images along the x, y, and z axes. 
\textbf{3D segmentation datasets:} BTCV~\cite{landman2015miccai}, MSD-Prostate~\cite{antonelli2022medical}, MSD-Spleen~\cite{antonelli2022medical}, MSD-Pancreas~\cite{antonelli2022medical}
.  In the cases of official test set absence, we divide the data as training and test sets in a 4:1 ratio. The detailed dataset information is summarized in Table \ref{combined_dataset}.

\begin{table}[t!]
\caption{Detail information for downstream medical image datasets with different tasks used in the experiment. ``Cla." is Classification and ``Seg." means Segmentation. Data size refers to the number of images.}
\setlength{\tabcolsep}{1.6pt}
\label{combined_dataset}
\begin{center}
\begin{tabular}{c|c|c|c|c}
\toprule[1.2pt]
 {\fontsize{9pt}{7pt}\selectfont Scenario} & {\fontsize{9pt}{7pt}\selectfont Dataset} &  {\fontsize{9pt}{7pt}\selectfont Targets}  &{\fontsize{9pt}{7pt}\selectfont Modality} & {\fontsize{9pt}{7pt}\selectfont Data Size} \\ 
 \midrule
 \multirow{5}{*}{2D Cla.} & ISIC2018~\cite{tschandl2018ISIC}  & Melanoma & RGB  & 11527    \\ 
  & COVID~\cite{xingyi2020covid_CT}   & COVID-19    & CT  & 746  \\ 
  & BTC~\cite{saleh2020BTC}   & Brain Tumor   & MRI & 3264    \\ 
  & BUSI~\cite{al2020busi}   & Breast Cancer    & Ultrasound   & 626   \\ 
  & ChestXray~\cite{cohen2020covid_xray}  &  Lung Diseases    & X-Ray  & 7135  \\ 
 \midrule
  \multirow{2}{*}{3D Cla.} & MMD~\cite{morozov2020mosmeddata}  & COVID-19 & CT  & 1110    \\ 
  & LAC~\cite{LAC2020}   & Lung Tumor   & CT  & 1050 \\ 
\midrule 
 \multirow{7}{*}{2D Seg.} & CDPRD~\cite{humans_in_the_loop_2023}  & Teeth & X-ray& 596 \\
  & FIVES~\cite{jin2022fives} & Retinal Vessels & Fundus & 698 \\
  & USC~\cite{USC2013} & Skin Cancer & Dermoscopy & 206 \\
  & TN3K~\cite{gong2021multi} & Thyroid Nodule & Ultrasound & 3400 \\
  & Robo~\cite{garcia2021image} & Surgical Tools & RGB & 500 \\
& CHAOS~\cite{kavur2019,CHAOS2021,CHAOSdata2019} & Abdominal Organs & MRI & 1240 \\
& CONIC~\cite{graham2024conic,graham2021lizard} & Cell Nucleus & RGB & 4841 \\
 \midrule
  \multirow{4}{*}{3D Seg.} 
  & MSD-Prostate~\cite{antonelli2022medical} & Prostate & MRI & 64 \\
  & MSD-Pancreas~\cite{antonelli2022medical} & Pancreas Tumor & CT & 281 \\
  & MSD-Spleen~\cite{antonelli2022medical} & Spleen & CT & 41 \\
  & BTCV~\cite{landman2015miccai} & Abdominal Organs & CT & 30 \\
 \bottomrule[1.2pt]
\end{tabular}
\end{center}
\end{table}

\subsection{Evaluation Metrics}
Two evaluation metrics were used to assess the performance of the medical image classification and segmentation models. For classification, accuracy was employed as a measure of correct predictions. For the segmentation task, the Dice Similarity Coefficient (DSC) was used to evaluate the overlap between the predicted segmentation and the ground truth, with values closer to 1 indicating better segmentation performance.

\subsection{Baselines}
We compared DRD with several baselines. First, we included the performance of medical foundation models and the vanilla-trained downstream models. 
For the absence of task-specific classification heads in foundation models, we employed full fine-tuning with a tailored classification head for 2D cases. For 3D scenarios, 
%
%we utilized a linear probe to assess the model's performance.
\rev{we employed full fine-tuning to assess the teacher’s performance.}
We also compared our method with several existing downstream adaptation techniques. For knowledge distillation, we evaluated Hint~\cite{romero2014fitnets}, AT~\cite{zagoruyko2016paying}, VID~\cite{ahn2019variational}, PKT~\cite{passalis2020probabilistic}, SemCKD~\cite{chen2021cross}, CRD~\cite{tian2019contrastive}, MGD~\cite{yang2022masked}, Norm~\cite{liu2023norm}, \rev{and OFA~\cite{hao2024one}, a representative heterogeneous distillation method.}
In addition, we included comparisons with various PEFT methods, such as the adapter tuning method AdaptFormer~\cite{chen2022adaptformer}, LoRA-based method MeLo~\cite{zhu2023melo}, prompt tuning methods VPT~\cite{jia2022visual} and LPT~\cite{dong2022lpt}, and the prefix tuning method VQT~\cite{tu2023visual}. 
% For a fair comparison, these methods also employ knowledge distillation to obtain the same size student models.
We also evaluated specialized adaptation methods designed for specific tasks, including the segmentation-specific distillation method CIRKD~\cite{yang2022cross} and MobileSAM~\cite{zhang2023faster}.
\subsection{Medical Foundation Models and Downstream Models}

\par{\noindent \bf Classification Scenarios.} 
We selected three 2D medical classification foundation models to validate the proposed method's effectiveness: PMC-CLIP~\cite{lin2023pmc}, RadDenseNet~\cite{mei2022radimagenet}, and LVM-Med~\cite{nguyen2023lvm}. 
PMC-CLIP is trained on 1.6 million pairs of image-caption data by contrastive learning with ResNet-50.
% , with a visual module structure of ResNet-50. 
RadDenseNet is trained on 1 million images by supervised learning with DenseNet-121. 
LVM-Med is trained on 1.3 million images by self-supervised learning with VIT-B. For the 2D downstream  models, we used three lightweight structures, namely ResNet18~\cite{he2016deep}, MobileNet~\cite{sandler2018mobilenetv2}, and ShuffleNet~\cite{ma2018shufflenet}.
Additionally, we extended our setup to 3D classification using the Merlin~\cite{blankemeier2024merlin} as the medical foundation model, trained on 6.3 million images, 1.8 million EHR diagnosis codes, and 6 million radiology report tokens. For downstream models, we used the 3D versions of ResNet18 and DenseNet121~\cite{hara2017learning}.

\par{\noindent \bf Segmentation Scenarios.}
We selected three foundation models for 2D segmentation: MedSAM~\cite{ma2024segment}, MSA~\cite{wu2023medical}, and Swin-UMamba~\cite{liu2024swin}, and one model for 3D segmentation: SAT~\cite{zhao2023one}. MedSAM, based on the SAM architecture, was adapted for medical images using 1.57 million image-mask pairs spanning 10 modalities and over 30 cancer types. MSA (also based on the SAM model) provides a variety of pre-trained checkpoints for different organs and lesions. Swin-UMamba is built on the emerging Mamba architecture tailored for medical use cases. SAT, trained on 22,000 3D medical images across 72 segmentation datasets, covers diverse anatomical sites, diseases, and modalities.
% SAT is distinguished by its training on an extensive collection of 22,000 3D medical images, spanning 72 segmentation datasets that cover various anatomical sites, diseases, and modalities.
% Downstream models need to be selected correspondingly due to the difference from foundation models. 
% For MedSAM and MSA, which accept not only image-mask pairs but also prompt inputs such as points or boxes, it is necessary to design downstream student models specifically tailored to match the input style of these foundation models. 
For MedSAM and MSA, which accept image-mask pairs and prompt inputs like points or boxes, downstream models should be designed to match their input style, with similar prompts used for consistency.
Inspired by MobileSAM~\cite{zhang2023faster}, we opted to use lightweight models such as ResNet18, ShuffleNet, MobileNet, and ViT-Tiny~\cite{touvron2021training} to replace the image encoder of SAM, while retaining the structure of the mask decoder and prompt encoder parts. 
\rev{For fair comparison, all downstream adaptation methods applied to SAM-like models use the same single bounding-box prompt for each sample, without any prompt sampling strategy.}
For the Swin-UMamba model, we selected lightweight segmentation models Unet and Unet3+~\cite{huang2020unet}. For the 3D segmentation tasks, we exclusively chose the classic 3D Unet segmentation model as the downstream model.

\rev{For both classification and segmentation scenarios, stage partitioning is defined coarsely in an architecture-aware manner: hierarchical CNN-like backbones follow natural stage boundaries, while flat transformer- or Mamba-like backbones are partitioned by grouping consecutive blocks. The teacher and student only need to share the same number of coarse stages, rather than the same internal layer counts.}

\subsection{Implementation Details}
We utilized the AdamW optimizer with a learning rate of 5e-3 for all experiments. For 2D classification tasks, each downstream dataset was run for 240 epochs with a batch size of 32. For 2D segmentation tasks, the models were trained for 100 epochs with a batch size of 8. Due to the high computational cost associated with processing 3D images, each 3D dataset was run for 100 epochs with a reduced batch size of 4. 
The input image sizes for training were set according to the foundation model’s requirements: 224×224 for LVM-Med, PMC-CLIP, and Rad-DenseNet; 224x224x160 for Merlin; 256×256 for Swin-UMamba; 1024×1024 for MedSAM and MSA; and 144×144×48 for SAT. 
In our loss function, the hyperparameters $\alpha$ and $\beta$ were initially set to 1 and decreased linearly throughout the training, so that the training gets more guidance at the early stage, but then progressively follows more relaxed constraints and the downstream supervision.
% For all experiments, we conducted them on the RTX 4090 GPUs.
\rev{All adaptation experiments were conducted on RTX 4090 GPUs. Due to the high memory cost, the full fine-tuning of the 3D teacher was conducted on a single 80GB NVIDIA A100.}

\begin{table*}[t!]%\footnotesize
\caption{{Comparison with KD methods in 2D and 3D classification experiments. The numbers next to the teacher model names represent their performance.
% Due to space constraints, we only display all the results on the BUSI and BTC here.
}}
\centering
\resizebox{0.97\textwidth}{!}{
\setlength{\tabcolsep}{2.6mm}{
\begin{tabular}{c|c|c|cccccccc|c}
\toprule[1.2pt]
\multicolumn{12}{c}{Downstream Task Adaptation for \textbf{2D Medical Classification} Scenarios} \\
\midrule[1.2pt]
 Dataset & Teacher & Student  &Vanilla & VID & SemCKD & Crd & Hint & MGD & Norm & \rev{OFA} & Ours  \\
\midrule
 \multirow{9}{*}{BUSI} 
 & \multirow{3}{*}{PMC-CLIP 88.31} 
   & ResNet18 &77.92 & 77.92 & 68.83 & 75.32 & 77.92 & 77.27 & 73.38 & \rev{83.12} & \textbf{88.31} \\
 && ShuffleNet &81.82 & 85.06 & 75.97 & 82.47 & 81.17 & 84.41 & 64.94 & \rev{72.08} & \textbf{86.36} \\
 && MobileNet &80.52 & 85.06 & 85.06 & 81.82 & 83.12 & 85.06 & 82.47 & \rev{77.27} & \textbf{86.36} \\
\cmidrule(r){2-12}
 & \multirow{3}{*}{RadImageNet 86.36}
   & ResNet18 &77.92 & 76.62 & 70.13 & 75.97 & 77.27 & 78.57 & 68.83 & \rev{76.62} & \textbf{88.31} \\
 && ShuffleNet &81.82 & 81.82 & 76.62 & 81.17 & 83.77 & 84.41 & 56.49 & \rev{61.69} & \textbf{85.71} \\
 && MobileNet &80.52 & 83.12 & 84.42 & 81.17 & 83.12 & 82.46 & 61.04 & \rev{73.38} & \textbf{87.01} \\
\cmidrule(r){2-12}
 & \multirow{3}{*}{LVM-Med 94.16}
   & ResNet18 &77.92 & 73.38 & 69.48 & 68.83 & 79.22 & 66.23 & 64.29 & \rev{80.52} & \textbf{88.96} \\
 && ShuffleNet &81.82 & 81.82 & 80.52 & 79.87 & 81.17 & 82.47 & 79.87 & \rev{75.32} & \textbf{85.71} \\
 && MobileNet &80.52 & 83.77 & 86.36 & 83.12 & 83.77 & 81.17 & 71.43 & \rev{79.22} & \textbf{87.01} \\
\midrule
% \midrule
 \multirow{9}{*}{BTC} 
 & \multirow{3}{*}{PMC-CLIP 80.20} 
   & ResNet18 &77.92 & 77.16 & 78.17 & 68.78 & 77.66 & 75.89 & 77.66 & \rev{72.59} & \textbf{80.71} \\
 && ShuffleNet &78.17 & 77.41 & 77.16 & 69.29 & 77.16 & 76.90 & 75.13 & \rev{69.29} & \textbf{79.44} \\
 && MobileNet &76.90 & 77.92 & 77.41 & 61.93 & 76.65 & 77.92 & 74.62 & \rev{71.07} & \textbf{79.70} \\
\cmidrule(r){2-12}
 & \multirow{3}{*}{RadImageNet 80.46}
   & ResNet18 &77.92 & 77.66 & 77.41 & 76.40 & 77.16 & 76.90 & 60.41 & \rev{68.53} & \textbf{78.68} \\
 && ShuffleNet &78.17 & 77.41 & 77.16 & 76.90 & 75.63 & 76.65 & 55.08 & \rev{65.99} & \textbf{79.95} \\
 && MobileNet &76.90 & 78.17 & 77.41 & 76.90 & 76.90 & 77.16 & 54.82 & \rev{73.35} & \textbf{79.95} \\
\cmidrule(r){2-12}
 & \multirow{3}{*}{LVM-Med 80.71}
   & ResNet18 &77.92 & 77.41 & 78.43 & 75.13 & 80.71 & 55.33 & 71.57 & \rev{70.30} & \textbf{81.98} \\
 && ShuffleNet &78.17 & 76.65 & 78.68 & 75.13 & 77.92 & 79.95 & 67.77 & \rev{67.77} & \textbf{80.46} \\
 && MobileNet &76.90 & 77.41 & 79.44 & 74.87 & 77.92 & 72.91 & 69.04 & \rev{70.81} & \textbf{82.49} \\
\midrule
 \multirow{3}{*}{ISIC} & PMC-CLIP 76.92 & \multirow{3}{*}{ResNet18}&70.50 & 75.73 & 75.66 & 76.26 & 75.33 & 77.05 & 75.79 & \rev{76.92} & \textbf{78.11} \\
 & RadImageNet 73.41 & &70.50& 75.99 & 75.99 & 76.32 & 76.52 & 77.18 & 63.76 & \rev{74.40} & \textbf{77.18} \\
 & LVM-Med 82.74 & &70.50& 75.93 & 77.18 & 78.77 & 79.23 & 75.99 & 78.37 & \rev{76.72} & \textbf{79.76} \\
\midrule
 \multirow{3}{*}{COVID} & PMC-CLIP 78.82 & \multirow{3}{*}{ResNet18}  &75.86 & 77.34 & 76.85 & 74.88 & 75.37 & 75.86 & 79.80 & \rev{80.30} & \textbf{81.28} \\
 & RadImageNet 77.34 & &75.86& 77.83 & 77.83 & 73.89 & 77.83 & 79.31 & 73.89 & \rev{\textbf{78.82}} & \textbf{78.82} \\
 & LVM-Med 83.74 & &75.86& 76.85 & 80.79 & 75.86 & 79.80 & 73.40 & 82.27 & \rev{79.80} & \textbf{82.27} \\
\midrule
 \multirow{3}{*}{ChestXray} & PMC-CLIP 96.11 & \multirow{3}{*}{ResNet18} &92.87 & 92.74 & 94.16 & 93.64 & 93.13 & 93.51 & 72.24 & \rev{-} & \textbf{95.33} \\
 & RadImageNet 94.55 & &92.87& 93.77 & 92.35 & 94.03 & 93.39 & 92.09 & 90.53 & \rev{-} & \textbf{94.68} \\
 & LVM-Med 94.16 & &92.87& 93.64 & 93.51 & 93.51 & 94.42 & 92.22 & 81.84 & \rev{-} & \textbf{96.11} \\
 \bottomrule[1.2pt]
\end{tabular}}}
\label{table_kd_new}
\vspace{-0.07cm}
\end{table*}

\begin{table*}[ht!]
\vspace{-0.2cm}
\centering
\resizebox{0.98\textwidth}{!}{
\setlength{\tabcolsep}{14pt}{
\begin{tabular}{c|c|c|ccccc|c}
\multicolumn{9}{c}{{Downstream Task Adaptation for \textbf{3D Medical Classification} Scenarios}} \\
\midrule[1.2pt]
Dataset & Teacher & Student & Vanilla & VID & SemCKD & PKT & Hint & Ours \\
\midrule
% \multirow{2}[2]{*}{MMD} & \multirow{2}[2]{*}{Merlin} & Resnet18 & 55.14  & 58.92  & 53.51  & 57.30  & 60.00  &- &- & 62.70  \\
%       &       & Densenet121 & 54.59  & 57.84  & 58.38  & 59.46  & 56.22  &- &- & 63.78  \\
% \midrule
% \multirow{2}[2]{*}{LAC} & \multirow{2}[2]{*}{Merlin} & Resnet18 & 54.29  & 58.86  & 54.29  & 57.14  & 58.29  &- &- & 60.57  \\
%       &       & Densenet121 & 63.43  & 61.71  & 63.43  & 66.29  & 64.57 &- &- & 69.71  \\ 
\multirow{2}[2]{*}{MMD} & \multirow{2}[2]{*}{Merlin \rev{74.64}} & Resnet3D & 55.14 & 60.00 & 57.30 & 53.51 & 58.92 & \textbf{62.70} \\
      &       & Densenet3D & 54.59 &56.22  & 59.46 & 58.38 & 57.84 & \textbf{63.78} \\
\midrule
\multirow{2}[2]{*}{LAC} & \multirow{2}[2]{*}{Merlin \rev{77.92}} & Resnet3D & 54.29 &58.29  & 57.14 &54.29  &58.86  & \textbf{60.57} \\
      &       & Densenet3D & 63.43 &64.57  & 66.29 & 63.43 & 61.71  & \textbf{69.71} \\

\bottomrule[1.2pt]
\end{tabular}
}}
\vspace{-7pt}
\end{table*}

\begin{table*}[ht!]
  \centering
  \caption{Comparing DRD with baselines across multiple datasets and medical foundation models. The numbers next to the teacher model names represent their performance. ResNet, ShuffleNet, MobileNet, ViT-Tiny here to the lightweight SAM-like models that replace the SAM image encoder with these models.}
    \resizebox{0.95\textwidth}{!}{
    \setlength{\tabcolsep}{7pt}{
    \begin{tabular}{c|c|c|ccccccc|c}
    \toprule[1.2pt]
    \multicolumn{11}{c}{Downstream Task Adaptation for \textbf{2D Medical Segmentation} Scenarios} \\
    \midrule[1.2pt] 
    Dataset & Teacher & Student & Vanilla & Hint & VID & SemCKD & PKT & CIRKD & MobSAM & Ours \\
    \midrule
    \multirow{8}[2]{*}{\makecell[c]{FIVES}} & \multirow{4}[2]{*}{MedSAM 16.53} & ResNet18 & 68.00 & 63.95  & 65.51  & 58.09  & 68.05  & 72.57  & 59.62  & \textbf{74.39} \\
          &       & ShuffleNet & 56.25 & 56.30  & 54.14  & 47.95  & 53.43  & 66.97  & 48.46  & \textbf{73.06} \\
          &       & MobileNet& 69.58  & 65.29  & 68.22  & 59.62  & 69.61  & 70.04  & 57.80  & \textbf{74.90} \\
          &       & ViT-Tiny &83.06    & 85.30  & 86.38  & 81.69  & 86.87  & 87.43  & 85.18  & \textbf{88.35} \\
    \cmidrule(r){2-11}
          & \multirow{4}[2]{*}{MSA 42.05} & ResNet18 &68.00  & 69.91  & 70.02  & 67.31  & 70.57  & 70.61  & 52.78  & \textbf{74.57} \\
          &       & ShuffleNet &56.25 & 60.46  & 59.62  & 57.60  & 57.10  & 70.29  & 46.01  & \textbf{72.83} \\
          &       & MobileNet &69.58  & 66.03  & 66.13  & 63.69  & 66.30  & 73.18  & 56.01  & \textbf{75.25} \\
          &       & ViT-Tiny &83.06    & 86.30  & 86.66  & 85.00  & 86.99  & 87.36  & 69.07  & \textbf{88.51} \\
    \midrule
    \multirow{8}[2]{*}{\makecell[c]{USC}} & \multirow{4}[2]{*}{MedSAM 83.73} & ResNet18 &85.32  & 88.93  & 88.07  & 89.15  & 89.32  & 88.96  & 88.77  & \textbf{90.63} \\
          &       & ShuffleNet &86.19 & 86.38  & 86.64  & 86.55  & 86.54  & 86.65  & \textbf{87.34} & 86.79 \\
          &       & MobileNet &86.47  & 86.65  & 88.44  & 86.82  & 86.87  & 87.46  & 87.61  & \textbf{88.68} \\
          &       & ViT-Tiny &87.93    & \textbf{90.22} & 89.04  & 89.88  & 88.26  & 88.59  & 89.86  & 89.23 \\
    \cmidrule(r){2-11}
          & \multirow{4}[2]{*} {MSA 87.88} & ResNet18 &85.32  & 88.37  & 89.69  & 88.60  & \textbf{89.97} & 88.34  & 82.74  & 89.42 \\
          &       & ShuffleNet &86.19 & 87.92  & 87.27  & 88.33  & 88.29  & 85.87  & 76.68  & \textbf{89.18} \\
          &       & MobileNet &88.47  & 87.90  & 87.62  & 87.92  & 87.08  & 85.04  & 87.60  & \textbf{89.44} \\
          &       & ViT-Tiny &87.93    & 88.72  & 88.95  & 88.73  & 89.39  & 88.05  & 62.78  & \textbf{90.81} \\
    \midrule
    \multirow{4}[2]{*}{CDPRD} & \multirow{4}[2]{*}{MedSAM 78.34} & ResNet18 &87.61  & 88.77  & 87.83  & 80.45  & 81.62  & 88.40  & 81.82  & \textbf{89.33} \\
          &       & ShuffleNet &76.54 & 79.92  & 77.28  & 78.83  & 76.49  & 81.66  & 77.54  & \textbf{87.54} \\
          &       & MobileNet &84.04  & 87.15  & 84.04  & 84.21  & 83.47  & 84.43  & 82.97  & \textbf{89.41} \\
          &       & ViT-Tiny &90.91    & 94.66  & 94.53  & 94.38  & 93.80  & 94.08  & 94.62  & \textbf{95.33} \\
    \midrule

    \multirow{4}[2]{*}{TN3K} & \multirow{4}[2]{*}{MSA 56.32} & ResNet18 &82.36  & 87.00  & 86.46  & 87.07  & 85.06  & 85.39  & 81.04  & \textbf{87.73} \\
          &       & ShuffleNet &82.32 & 84.84  & 85.50  & 84.85  & 84.51  & 84.31  & 83.59  & \textbf{86.99} \\
          &       & MobileNet &83.29  & 87.44  & 86.24  & 84.91  & 85.56  & 83.10  & 86.06  & \textbf{88.16} \\
          &       & ViT-Tiny &84.96    & 84.29  & 84.44  & 84.66  & 83.68  & 85.47  & 80.53  & \textbf{86.92} \\
    \midrule
    \multirow{2}[2]{*}{CONIC} & \multicolumn{1}{c|}{\multirow{2}[2]{*}{Swin-UM 83.02}} & UNet &82.87 & 82.74  & 83.06  & 82.76  & 83.10  & 83.06  & - & \textbf{84.83} \\
          &       & UNet3+ &80.39 & 82.43  & 80.07  & 81.71  & 80.70  & 80.51 & - & \textbf{83.27} \\
    \midrule
    \multirow{2}[2]{*}{Robo} & \multicolumn{1}{c|}{\multirow{2}[2]{*}{Swin-UM 81.60}} & UNet &82.47 & 88.17  & 86.69  & 88.35  & 88.35  & 87.72 & - & \textbf{88.76} \\
          &       & UNet3+ &83.80 & 84.54  & 83.97  & 85.19  & 83.24  & 81.13  & - & \textbf{85.53} \\
    \midrule
    \multirow{2}[2]{*}{CHAOS} & \multicolumn{1}{c|}{\multirow{2}[2]{*}{Swin-UM 84.51}} & UNet &90.04 & 93.68  & 93.36  & 93.20  & 93.78  & 92.94 & - & \textbf{94.51} \\
          &       & UNet3+ &90.70 & 93.16  & 93.70  & 92.84  & 93.07  & 93.03 & - & \textbf{94.05} \\
    % \midrule[1.2pt]
    \end{tabular}}}
    \vspace{-0.15cm}
  \label{tab:seg_results}
\end{table*} 
\begin{table*}[ht!]
\vspace{-0.2cm}
\centering
    \resizebox{0.95\textwidth}{!}{
    \setlength{\tabcolsep}{11pt}{
    \begin{tabular}{c|c|c|ccccc|c}
    \toprule[1.2pt]
    \multicolumn{9}{c}{Downstream Task Adaptation for \textbf{3D Medical Segmentation} Scenarios} \\
    \midrule[1.2pt] 
    Dataset & Teacher & Student & Vanilla & Hint & VID & SemCKD & PKT  & Ours \\
    \midrule
    % BTCV  & SAT 52.25 & Unet3D & 42.49 & 49.16 & 45.28 & 48.73 & 45.91 & -    & -    & \textbf{58.87} \\
    BTCV-Stomach & SAT 21.57  & Unet3D  & 57.90  & 62.48  & 59.39  & 59.79  & 58.61  & \textbf{66.47}  \\
    BTCV-kidney & SAT 64.37  & Unet3D  & 50.43  & 57.76  & 54.82  & 64.72  & 58.85  & \textbf{66.33}  \\
    BTCV-Gallbladder &SAT 70.80  & Unet3D  & 19.15  & 27.23  & 21.64  & 21.67  & 20.27  & \textbf{43.80}  \\  
    \midrule
    MSD-Spleen & SAT 87.21  & Unet3D & 80.87  & 87.78  & 86.88  & 85.75  & 87.15     & \textbf{89.83}  \\
    \midrule
    MSD-Prostate & SAT 72.76  & Unet3D & 75.18  & 77.46  & 77.11  & 76.36  & 76.52    & \textbf{77.55}  \\
    \midrule
    MSD-Pancreas & SAT 77.21  & Unet3D & 20.07  & 65.88  & 65.83  & 64.89  & 64.70    & \textbf{66.41}  \\
    \bottomrule[1.2pt]
    \end{tabular}}}
\vspace{-7pt}
\end{table*}

\begin{figure*}[t!]
\centerline{\includegraphics[width=0.97\linewidth]{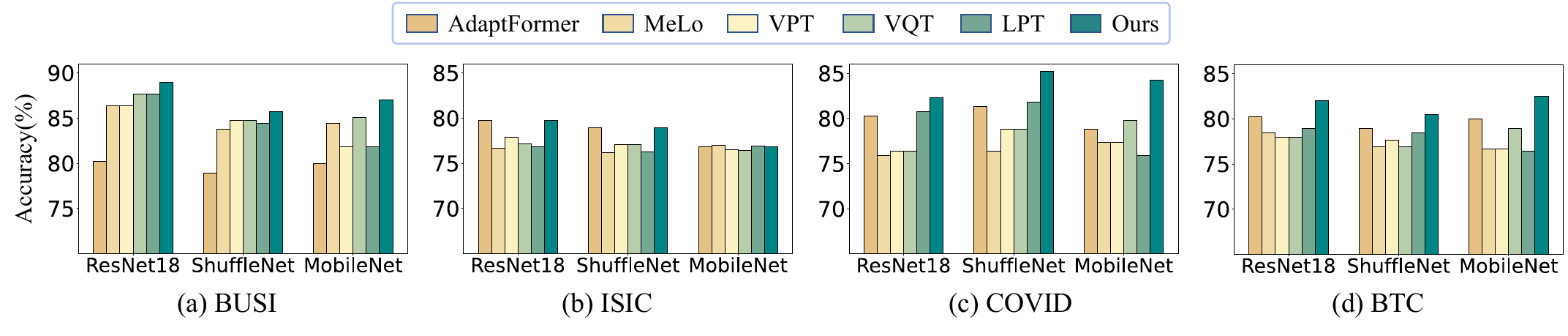}}
\caption{Comparison with PEFT methods in classification experiments. Subfigures (a)-(d) present the results of four datasets respectively. Since PEFT is mainly applied to transformer structures, the foundation model here is LVM-Med (ViT-Based). }
\label{fig:PEFT_results}
\end{figure*}

\section{Results}
In this section, we provide the comprehensive experimental verification of our method and in-depth analysis.

\subsection{Main Results on Classification}
In this subsection, we present the classification results of DRD against the relevant KD methods and PEFT methods.
\subsubsection{Compared with KD methods}
We compare DRD with other KD methods on 2D/3D Classification Downstream Tasks in Table~\ref{table_kd_new}.
As can be seen, DRD is superior overall to previous KD methods, even approaching the performance of fully fine-tuning 2D foundation models 
% and surpassing the linear probe performance of 3D foundation models, 
while using more lightweight structures.
The improvement offered by KD methods over directly training a student model is limited, and the performance enhancements vary significantly across different datasets. For example, when using the BUSI dataset, the results of almost all KD methods are inferior to those achieved by directly training student models. This indicates that direct knowledge distillation, without addressing task or structure inconsistencies, is inefficient and can sometimes be counterproductive to downstream model training.  
In contrast, DRD significantly enhances downstream model performance across various datasets, structures. Consequently, DRD proves to be a versatile and robust approach for improving model performance in diverse applications.

\subsubsection{Compared with PEFT methods}
Fig.~\ref{fig:PEFT_results} compares DRD with several Parameter-Efficient Fine-Tuning (PEFT) methods on various classification tasks. For a fair comparison, these methods also employ knowledge distillation to obtain the same student models. The foundation model used in these experiments is LVM-Med, as PEFT methods are primarily designed for transformer-based architectures.
The results indicate that the performance of PEFT methods varies considerably across different datasets. For instance, while the AdaptFormer method performs well on several datasets, its performance declines significantly on the BUSI.
However, DRD consistently proves a significant performance, particularly on smaller datasets such as BUSI and COVID, where it achieves the highest accuracy in nearly all configurations.
%当下游数据量足够大时，学生模型通常能够从大量样本中学习到足够的特征表示，获取足够丰富的知识，拉近与教师之间的差距。因此重编程模块（Reprogramming Module）可能未能对模型的特征表示做出显著改善，从而导致我们方法额外带来的增益不明显。但是在现实的医疗场景中，数据的收集开销巨大，获取的下游数据集难以达到较大的规模。修改一下。
For larger datasets like ISIC, DRD's performance is comparable to that of PEFT methods, which reflects that PEFT techniques benefit from a larger quantity of training data.
In contrast, our method consistently maintains strong performance, regardless of dataset size. This suggests that DRD is less dependent on large volumes of labeled data, which actually is critical to medical scenarios where data collection and labeling are expensive and time-consuming.

\subsection{Main Results on Segmentation}
In this subsection, we present the comparative results of DRD against baselines on segmentation foundation models.

\subsubsection{Experiments on MedSAM and MSA}  
% The results of comparing the foundation model using the SAM architecture with baselines for the downstream segmentation task are shown in Table~\ref{tab:seg_results}.
As shown in Table~\ref{tab:seg_results}, DRD method consistently outperformed established baselines across multiple medical imaging datasets and different downstream models. In some cases, DRD even significantly outperforms medical foundation models on downstream tasks with smaller deployment cost.
Notably, we observed that on the challenging FIVES dataset, both MedSAM and MSA exhibited poor zero-shot performance. However, DRD method demonstrated a consistently superior advantage compared to other baseline approaches across all student models, particularly when the student models had relatively worse performance. For instance, Shufflenet significantly underperforms relative to other student models, and while other baseline methods offer limited or even negative improvements, DRD substantially enhances its performance, achieving nearly a 17 percentage point increase over vanilla-trained model. This highlights the robustness of the DRD method when dealing with various student model architectures.
% , affirming its superior adaptability across different settings.

\subsubsection{Experiments on Swin-UMamba} 
In experiments of Swin-UMamba, we chose the near-domain datasets of its pre-training task as the downstream datasets. 
% Since we don't need to deal with prompt inputs like SAM architecture, we chose the models such as Unet and Unet3+ to be the downstream model.
From Table~\ref{tab:seg_results}, we can see that DRD exhibits strong adaptability, even when dealing with models based on entirely different architectures, such as Mamba based on state space models, and the UNet that utilizes a convolutional neural network framework. DRD consistently enhances downstream model performance and surpasses other adaptation methods, demonstrating the merit of reprogramming the teacher model's features across architectures into highly compatible downstream model space.

% \begin{figure*}[t]
%     \centering
%     \begin{minipage}{0.49\textwidth}
%         {\includegraphics[width=\textwidth]{img/cost_cla.png}}
%     \end{minipage}
%     \hfill % 添加水平间距
%     \begin{minipage}{0.49\textwidth}
%         {\includegraphics[width=\textwidth]{img/cost_seg.png}}
%     \end{minipage}

%     \caption{Comparison of cost and performance of models. The left figure shows classification models, the right shows segmentation models. Dot size indicates model parameter counts; * denotes models using the DRD method.
%     }
%     \label{cost_compare}
% \end{figure*}

% \begin{figure}[htp]
%     \centering
%     \includegraphics[width=\linewidth]{img/SAT_final.pdf}
%     \caption{3D segmentation task results: SAT Foundation Model with Unet Downstream Model. Only the specific values for SAT and DRD are labelled in the graph. KD represents the average of knowledge distillation methods: Hint, VID, SemCKD, and PKT. Dashed line for large SAT deployment model, solid line for small Unet deployment model.}
%     \label{fig:SAT_result_new}
% \end{figure}

\subsubsection{Results on SAT} 
In the 3D segmentation task, we compared the performance of DRD with SAT, Unet without pre-training, and knowledge distillation methods.
It is evident that the DRD method significantly and consistently outperforms both the Unet and knowledge distillation approaches. In most cases, DRD enables Unet that has a lower deployment cost, to surpass the foundation model SAT in performance on downstream tasks. This is particularly evident in the BTCV-Stomach task, where we found that SAT performed poorly, likely due to the absence of similar labels in its pre-training data. Nevertheless, our method still achieved substantial improvements over the baseline, suggesting that even when specific knowledge for downstream tasks is lacking, DRD can extract general and effective knowledge from SAT. Conversely, in the BTCV-Gallbladder task, SAT exhibits performance that far exceeded the Unet. In this case, the knowledge distillation methods resulted in only a modest improvement, with an average enhancement of 3.5\%. In contrast, our method achieved a notable improvement of 24.7\%, further demonstrating the high efficiency of DRD in knowledge transfer.

%加一个模型在RD之前的特征与RD之后的特征与student的特征比较图

% \subsection{Ablation Studies}
\subsection{Further analysis}

\subsubsection{\rev{Multi-seed Robustness}}
\rev{To quantify robustness more explicitly, we repeated representative classification and segmentation settings with four random seeds. Specifically, we evaluate BUSI and BTC with LVM-Med as the teacher for classification, and FIVES and CDPRD with MedSAM as the teacher for segmentation. For compact presentation, Table~\ref{tab:multi_seed} reports only the strongest competing baseline and DRD in each setting. Across these 14 settings, DRD achieves the best mean performance in all cases. We further conduct paired two-sided t-tests by comparing DRD against the strongest competing baseline in each setting: 13 out of 14 comparisons are significant at $p<0.05$, while the remaining FIVES/ViT-Tiny case is not statistically significant. These results provide direct statistical evidence that DRD is both effective and stable across random seeds.}

\begin{table}[t!]
    \centering
    \caption{\rev{Four-seed results on representative adaptation settings. Only the strongest baseline is shown. Superscript $^{*}$ indicates $p<0.05$, and $^{\mathrm{ns}}$ denotes no significant difference.}}
    \setlength{\tabcolsep}{4pt}
    \renewcommand{\arraystretch}{1.04}
    \resizebox{0.92\columnwidth}{!}{%
    \rev{%
    \begin{tabular}{c|c|c|c}
        \toprule
        Dataset & Student & Best baseline & Ours \\
        \midrule
        \multicolumn{4}{c}{\textbf{Cla. (teacher: LVM-Med)}} \\
        \midrule
        \multirow{3}{*}{BUSI}
            & ResNet18 & 83.60 $\pm$ 3.05 (Hint) & \textbf{88.15 $\pm$ 0.85}$^{*}$ \\
            & ShuffleNet & 79.06 $\pm$ 0.97 (SemCKD) & \textbf{86.36 $\pm$ 1.00}$^{*}$ \\
            & MobileNet & 84.58 $\pm$ 1.83 (SemCKD) & \textbf{87.01 $\pm$ 0.43}$^{*}$ \\
        \cmidrule(lr){1-4}
        \multirow{3}{*}{BTC}
            & ResNet18 & 78.68 $\pm$ 1.21 (Hint) & \textbf{79.70 $\pm$ 1.38}$^{*}$ \\
            & ShuffleNet & 75.44 $\pm$ 2.18 (SemCKD) & \textbf{80.20 $\pm$ 0.57}$^{*}$ \\
            & MobileNet & 77.85 $\pm$ 1.16 (SemCKD) & \textbf{80.46 $\pm$ 1.52}$^{*}$ \\
        \midrule
        \multicolumn{4}{c}{\textbf{Seg. (teacher: MedSAM)}} \\
        \midrule
        \multirow{4}{*}{FIVES}
            & ResNet18 & 70.98 $\pm$ 1.08 (CIRKD) & \textbf{72.16 $\pm$ 1.49}$^{*}$ \\
            & ShuffleNet & 67.83 $\pm$ 0.61 (CIRKD) & \textbf{70.76 $\pm$ 1.54}$^{*}$ \\
            & MobileNet & 71.97 $\pm$ 1.30 (CIRKD) & \textbf{75.22 $\pm$ 0.23}$^{*}$ \\
            & ViT-Tiny & 87.02 $\pm$ 0.30 (PKT) & \textbf{88.65 $\pm$ 0.21}$^{\mathrm{ns}}$ \\
        \cmidrule(lr){1-4}
        \multirow{4}{*}{CDPRD}
            & ResNet18 & 85.09 $\pm$ 2.21 (CIRKD) & \textbf{86.74 $\pm$ 1.73}$^{*}$ \\
            & ShuffleNet & 82.09 $\pm$ 0.50 (CIRKD) & \textbf{85.28 $\pm$ 1.52}$^{*}$ \\
            & MobileNet & 87.09 $\pm$ 0.05 (Hint) & \textbf{89.85 $\pm$ 0.36}$^{*}$ \\
            & ViT-Tiny & 91.78 $\pm$ 1.92 (Hint) & \textbf{92.11 $\pm$ 1.15}$^{*}$ \\
        \bottomrule
    \end{tabular}}}%
    \label{tab:multi_seed}
    \vspace{-0.5cm}
\end{table}

\begin{figure*}[t!]
\centerline{\includegraphics[width=\linewidth]{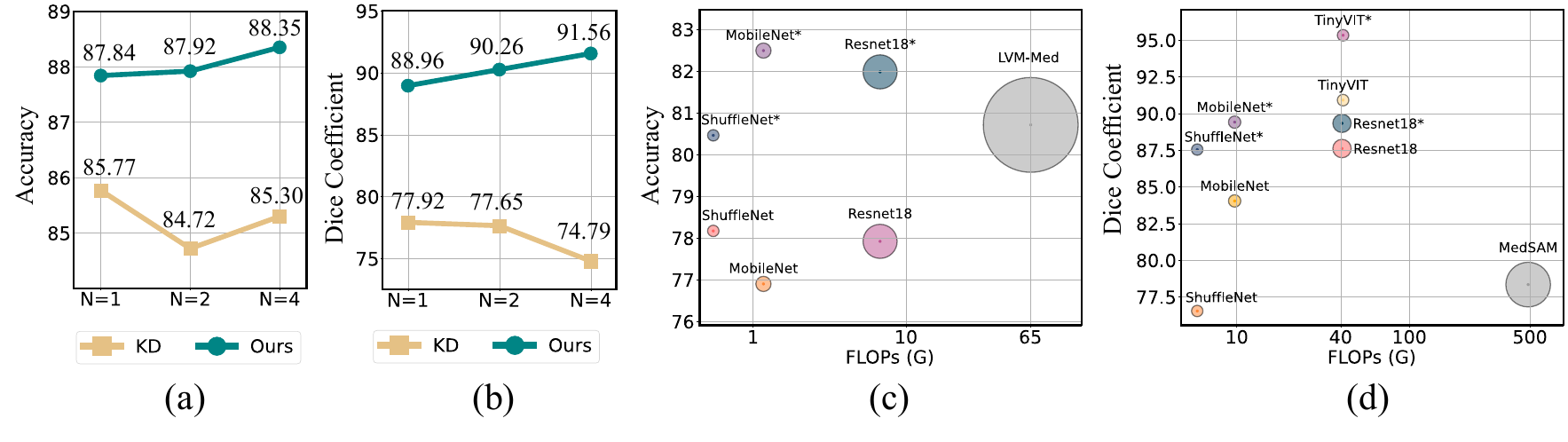}}
\vspace{-10pt}
\caption{Subfigures (a) and (b) compare the effects of varying depths on KD and DRD in classification and segmentation tasks, respectively. Subfigures (c) and (d) compare the cost and performance of models in classification and segmentation respectively, where the circular radius is proportional to the model parameter size, and the asterisk (*) means model using DRD.}
\label{fig:abla1+fu1}
\end{figure*}

%ablation 
\subsubsection{Ablation of different components}
We conduct the ablation experiments to illustrate the effectiveness of the DRD components by removing or replacing the key components and then conducting comparative experiments.
The results are shown in Table~\ref{tab:ablation_study}, where ``Di. Reprog'' and ``Co. Reprog'' represent direct reprogramming~\cite{chen2022model} and our co-training reprogramming respectively. 
It can be seen that replacing direct reprogramming with our co-training reprogramming results in a significant performance improvement. This indicates that the co-training reprogramming can better transfer knowledge to downstream models while mitigating task and model inconsistencies. Besides, the CKA distillation similarly improves performance, which could be attributed to robust knowledge transfer that tends to focus on the critical patterns in features, reducing the impact of randomness on training.

\begin{table}[t!]
    \centering
    \caption{Ablation of different components in DRD.}
    \begin{tabular}{cccc|c}
        \toprule
          \textbf{Di. Reprog.} & \textbf{Co. Reprog.} & \textbf{KD} & \textbf{CKA} & \textbf{Acc}  \\
        \midrule
          &  &  &  & 75.86 \\
        \midrule
         \cmark &  & \cmark &  & 80.79 \\
          & \cmark & \cmark &  & 82.27 \\
           & \cmark & \cmark & \cmark & \textbf{85.71} \\
        \bottomrule
    \end{tabular}
  \label{tab:ablation_study}%
\end{table}

% \begin{figure}[htb]
% \centerline{\includegraphics[width=0.97\linewidth]{img/ablav2.pdf}}
% \caption{Comparing experiments on the effect of depth in classification and segmentation scenarios. }
% \label{fig:ablation1}
% \end{figure}

\subsubsection{\rev{Ablation of Reprogramming Depth and Stability Analysis}}
In this part, we target to explore the effect of reprogramming depths in DRD and a baseline of KD to directly mimic features between the teacher and student models under that depth. 
Specifically, we divide the foundation model into $N$ segments, and when $N=1$, the model is treated as a black box, processing only the output features from the final layer. Similarly, when $N=2$ or $N=4$, the foundation model is divided into 2 or 4 segments, allowing for the extraction of intermediate features for transfer. We study the classification and segmentation performance of DRD and the baseline KD under different depths in Fig.~\ref{fig:abla1+fu1}(a) and Fig.~\ref{fig:abla1+fu1}(b). 
As can be seen, DRD significantly outperforms KD when N=1, indicating its strong knowledge transfer capability even when medical foundation models are treated as black box. As $N$ increases and features become deeper, the performance of KD fluctuates and declines, whereas our method consistently improves. This suggests that KD struggles to efficiently transfer deeper intermediate features from the foundation model. In contrast, DRD can effectively leverage deep features and ensures efficient knowledge transfer.

\rev{Beyond the performance trend, we further examine whether increasing $N$ affects optimization stability. As detailed in Appendix~\ref{app:conv_n} and Appendix~\ref{app:grad_diag}, DRD exhibits smooth convergence under different reprogramming depths ($N=1,2,3,4$), while the parameter-level diagnosis shows that the main auxiliary gradients remain aligned with the supervised gradient. These results suggest that using a larger reprogramming depth does not introduce severe convergence instability or destructive late-block gradient conflict.}

\subsubsection{\rev{Ablation of Reprogramming Block Design}}
\rev{Besides the reprogramming depth, we further ablate the design of the reprogramming projector on BUSI and BTC with LVM-Med as the teacher and ResNet18 as the student. We compare a linear mapping, a resize + $1\times1$ convolution, a 2-layer convolutional block, a wider 3-layer convolutional block, and our default design. Table~\ref{tab:block_ablation} shows that overly simple projectors underfit the cross-model transformation, while overly wide convolutional blocks increase cost substantially. Our default design achieves the best accuracy on both datasets, improving over the linear baseline by 2.70\% on BUSI and 3.17\% on BTC, while maintaining a much better accuracy-cost trade-off than the widest alternative.}

\subsubsection{\rev{Boundary Sensitivity and Stage Alignment}}
\begin{figure}[t]
\centering
\includegraphics[width=\linewidth]{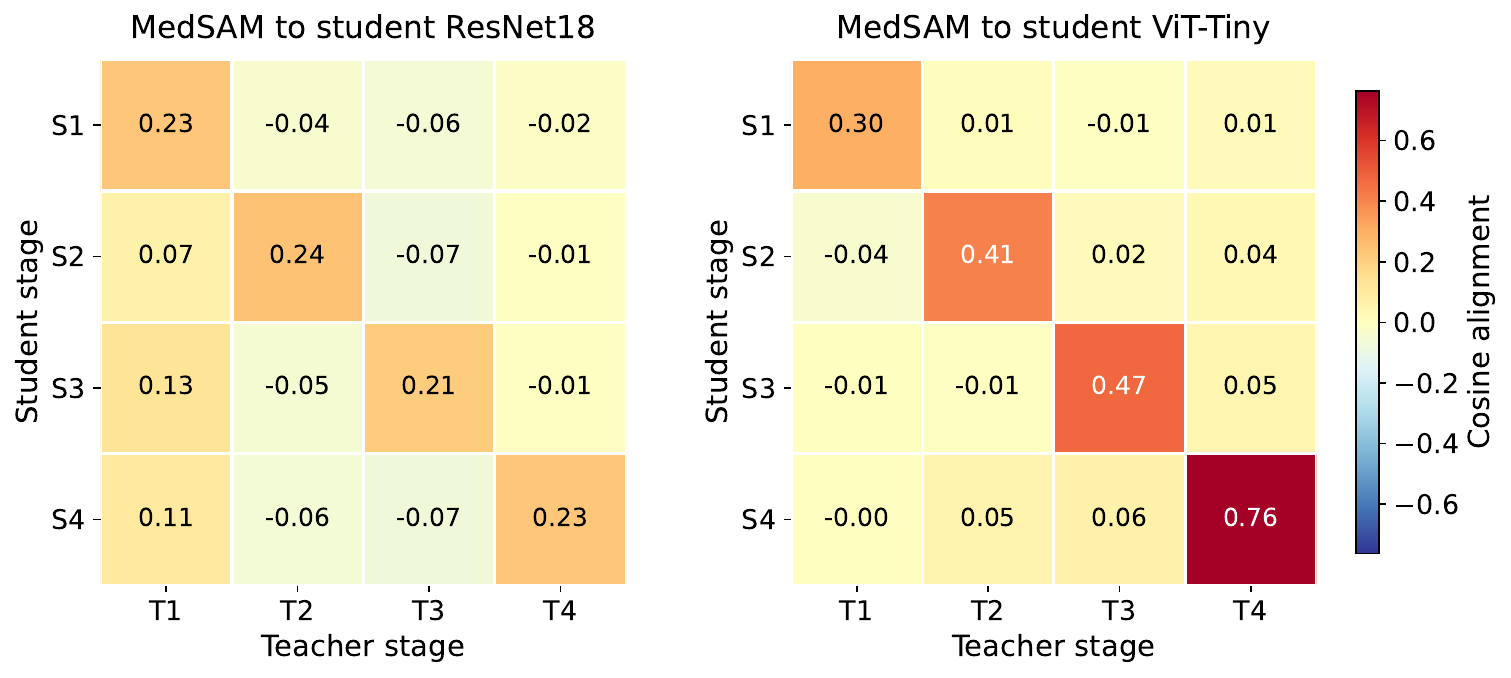}
\caption{\rev{Post-training cosine similarity matrices between teacher and student coarse stages. }}
\label{fig:reply_feature_alignment}
\vspace{-0.4cm}
\end{figure}

\rev{We further analyze whether DRD depends on a delicate manual partition of heterogeneous teacher and student models. Specifically, we vary the selected teacher output blocks used to define the four coarse stages while keeping the other settings unchanged. As shown in Table~\ref{tab:reply_boundary_pairing}~(a), the performance remains stable across reasonable boundary choices, with a fluctuation range of only 0.90\% for ResNet18 and 0.67\% for ViT-Tiny, indicating that DRD is not sensitive to a delicate hand-crafted boundary. 
We also test deliberately mismatched stage pairings under the same setting. In Table~\ref{tab:reply_boundary_pairing}~(b), compared with the normal identity pairing, reverse pairing drops by 5.02\% for ResNet18 and 4.67\% for ViT-Tiny, while shift-right pairing also degrades performance, showing that the transfer is not arbitrary and still benefits from stage-aware correspondence. 
Finally, Fig.~\ref{fig:reply_feature_alignment} presents the post-training cosine similarity matrices between teacher and student stages, where the diagonals are consistently brighter than the off-diagonal regions, suggesting that DRD learns an emergent stage-wise alignment during optimization rather than relying on pre-defined blockwise semantic equivalence.}

\begin{table}[t!]
    \centering
    \caption{\rev{Ablation of reprogramming block design.}}
    \resizebox{\linewidth}{!}{
    {\rev{
    \begin{tabular}{l|cc|cc}
        \toprule
        Projection block & BUSI Acc & BTC Acc & Params & GFLOPs \\
        \midrule
        Linear & 85.45 $\pm$ 0.61 & 76.53 $\pm$ 0.67 & 0.393M & 0.077 \\
        Resize + 1$\times$1 & 86.58 $\pm$ 1.33 & 77.51 $\pm$ 0.32 & 0.393M & 0.019 \\
        Conv-2 & 86.58 $\pm$ 0.61 & 78.53 $\pm$ 1.21 & 8.849M & 1.735 \\
        Wide Conv-3 & 87.01 $\pm$ 0.92 & 78.76 $\pm$ 0.32 & 23.206M & 4.549 \\
        Ours & \textbf{88.15 $\pm$ 0.85} & \textbf{79.70 $\pm$ 1.38} & 10.752M & 2.108 \\
        \bottomrule
    \end{tabular}}}}
    \label{tab:block_ablation}
\end{table}

\begin{table}[t!]
\centering
\caption{\rev{Boundary sensitivity and stage pairing ablations with MedSAM as teacher.}}
\label{tab:reply_boundary_pairing}
\setlength{\tabcolsep}{4pt}
\renewcommand{\arraystretch}{1.1}

\begingroup
\rev{
\begin{tabular}{l l c c}
\toprule
\multicolumn{4}{c}{\textbf{(a) Boundary sensitivity analysis.}}\\
\midrule
Student & Selected Blocks & Dice (\%) & $\Delta$ vs. Def. (\%) \\
\midrule
\multirow{5}{*}{ResNet18} 
    & 1-2-5-11        & 90.26 & -0.37 \\
    & 1-4-7-10        & 90.97 & +0.34 \\
    & 1-6-9-11        & 89.73 & -0.90 \\
    & 2-6-9-11        & 89.92 & -0.71 \\
    & 2-5-8-11 (Def.) & 90.63 & 0.00 \\
\midrule
\multirow{5}{*}{ViT-Tiny} 
    & 1-2-5-11        & 89.42 & +0.19 \\
    & 1-4-7-10        & 89.06 & -0.17 \\
    & 1-6-9-11        & 88.91 & -0.32 \\
    & 2-6-9-11        & 88.56 & -0.67 \\
    & 2-5-8-11 (Def.) & 89.23 & 0.00 \\
\midrule[0.08em]
\multicolumn{4}{c}{\textbf{(b) Stage pairing strategy ablation.}}\\
\midrule
Student & Pairing Strategy & Dice (\%) & $\Delta$ vs. Id. (\%) \\
\midrule
\multirow{3}{*}{ResNet18} 
    & Identity      & \textbf{90.63} & 0.00 \\
    & Reverse       & 85.61 & -5.02 \\
    & Shift-Right   & 86.70 & -3.93 \\
\midrule
\multirow{3}{*}{ViT-Tiny} 
    & Identity      & \textbf{89.23} & 0.00 \\
    & Reverse       & 84.56 & -4.67 \\
    & Shift-Right   & 87.28 & -1.95 \\
\bottomrule
\end{tabular}}
\endgroup

\end{table}

\subsubsection{Model Costs Comparison} 
In Fig.~\ref{fig:abla1+fu1}(c) and Fig.~\ref{fig:abla1+fu1}(d), we visualize the cost and performance of the models with or without DRD. 
It can be observed that DRD achieves performance comparable to that of the foundation model while significantly reducing the model parameter size and computational costs during inference. Additionally, when compared to a model trained using the vanilla method with the same deployment cost, DRD also shows a significant performance improvement.
It is worth noting that in 3D segmentation tasks, the significant computational cost of SAT hinders its deployment on consumer-grade GPUs, such as NVIDIA GeForce RTX 4090. However, by utilizing DRD, this overhead can be significantly reduced, which allows a lightweight model to be effectively deployed on the RTX 4090 while still achieving high performance in downstream tasks.

\begin{table}[t!]
\centering
\caption{\rev{Measured training overhead. Time is the average per-iteration wall-clock time, memory is peak GPU allocation, and Acc is four-seed mean/std. }}
\setlength{\tabcolsep}{4pt}
\resizebox{0.45\textwidth}{!}{
\rev{
\begin{tabular}{lccc}
\toprule
\textbf{Method} & \textbf{Time (ms/iter)} & \textbf{Memory (MB)} & \textbf{Acc} \\
\midrule
VID & 134.94 & 2768 & 75.89 $\pm$ 1.45 \\
OFA & 194.67 & 2670 & 70.94 $\pm$ 0.94 \\
LoRA & 235.32 & 9194 & 78.32 $\pm$ 1.97 \\
Full Fine-tuning & 271.94 & 11065 & 79.24 $\pm$ 1.96 \\
DRD w/o CKA & 141.31 & 4170 & 76.10 $\pm$ 2.45 \\
\textbf{DRD} & 150.29 & 4379 & \textbf{79.70 $\pm$ 1.38} \\
\bottomrule
\end{tabular}}}
\label{tab:overhead_compare}
\vspace{-0.25cm}
\end{table}

\rev{We further evaluate the training efficiency of DRD, including full fine-tuning of the foundation model as a reference. As summarized in Table~\ref{tab:overhead_compare}, all measurements were conducted on the BTC dataset under the LVM-Med-to-ResNet18 adaptation setting using a single NVIDIA RTX 4090 GPU.
Compared with full fine-tuning of the foundation model, DRD reduces the average per-iteration wall-clock time from 271.94 to 150.29 ms (\textbf{44.73\%}) and the peak GPU memory from 11065 to 4379 MB (\textbf{60.42\%}), while achieving comparable and slightly higher mean accuracy. 
Compared with the vanilla KD, DRD is moderately more expensive than VID during training (+15.35 ms/iter and +1611 MB), but improves the mean accuracy from 75.89 to 79.70; compared with OFA and LoRA, DRD is not only faster in wall-clock time, but also achieves higher mean accuracy (79.70 vs. 70.94/78.32).
To quantify the per-iteration cost of CKA, we compare DRD with its variant without CKA. Adding CKA increases the average per-iteration wall-clock time by only 6.35\% and the peak GPU memory by 5.01\%, while improving the mean accuracy by 3.60 percentage points, corresponding to a relative gain of 4.73\%. It also reduces the standard deviation by 43.67\%, suggesting that CKA provides a favorable efficiency--robustness trade-off.}

\begin{figure*}[t!]
\centerline{\includegraphics[width=0.92\linewidth]{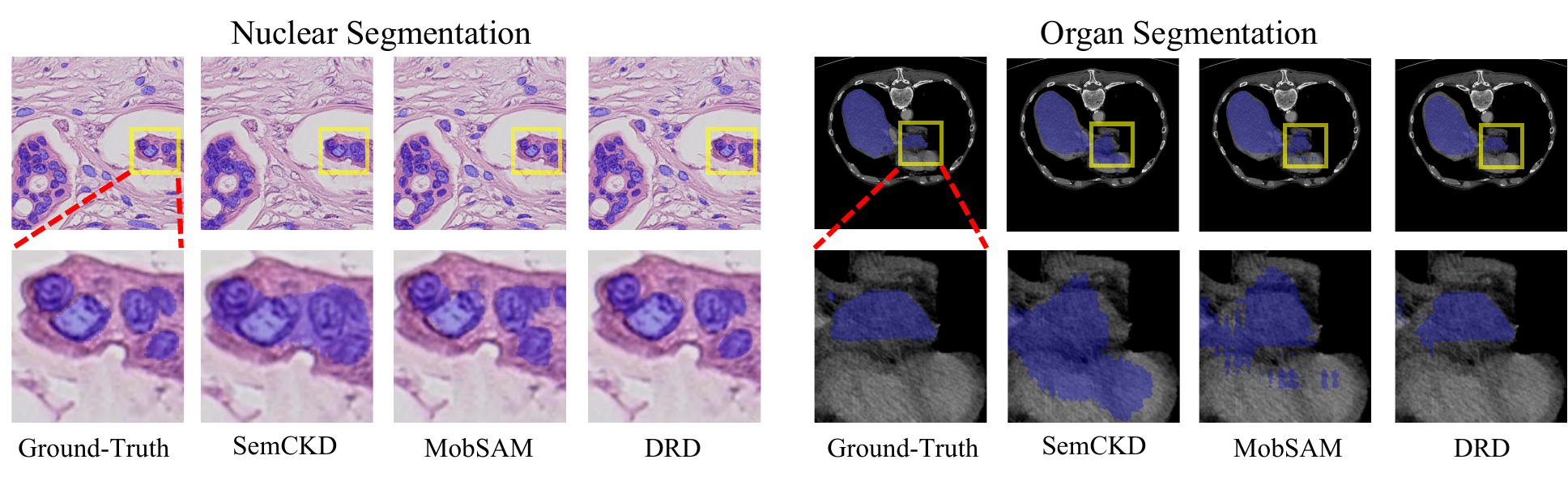}}
\vspace{-7pt}
\caption{Comparison of segmentation results. Randomly selected images were input into different models for nuclear and organ segmentation, and the results were compared with the ground truth. Blue masks represent the segmentation targets. The second-row images are enlarged versions of the yellow-boxed regions in the first-row images.}
\label{fig：case_study}
\vspace{-7pt}
\end{figure*}

\begin{figure}[t]
    \centering
    \begin{minipage}{0.24\textwidth}
        {\includegraphics[width=\textwidth]{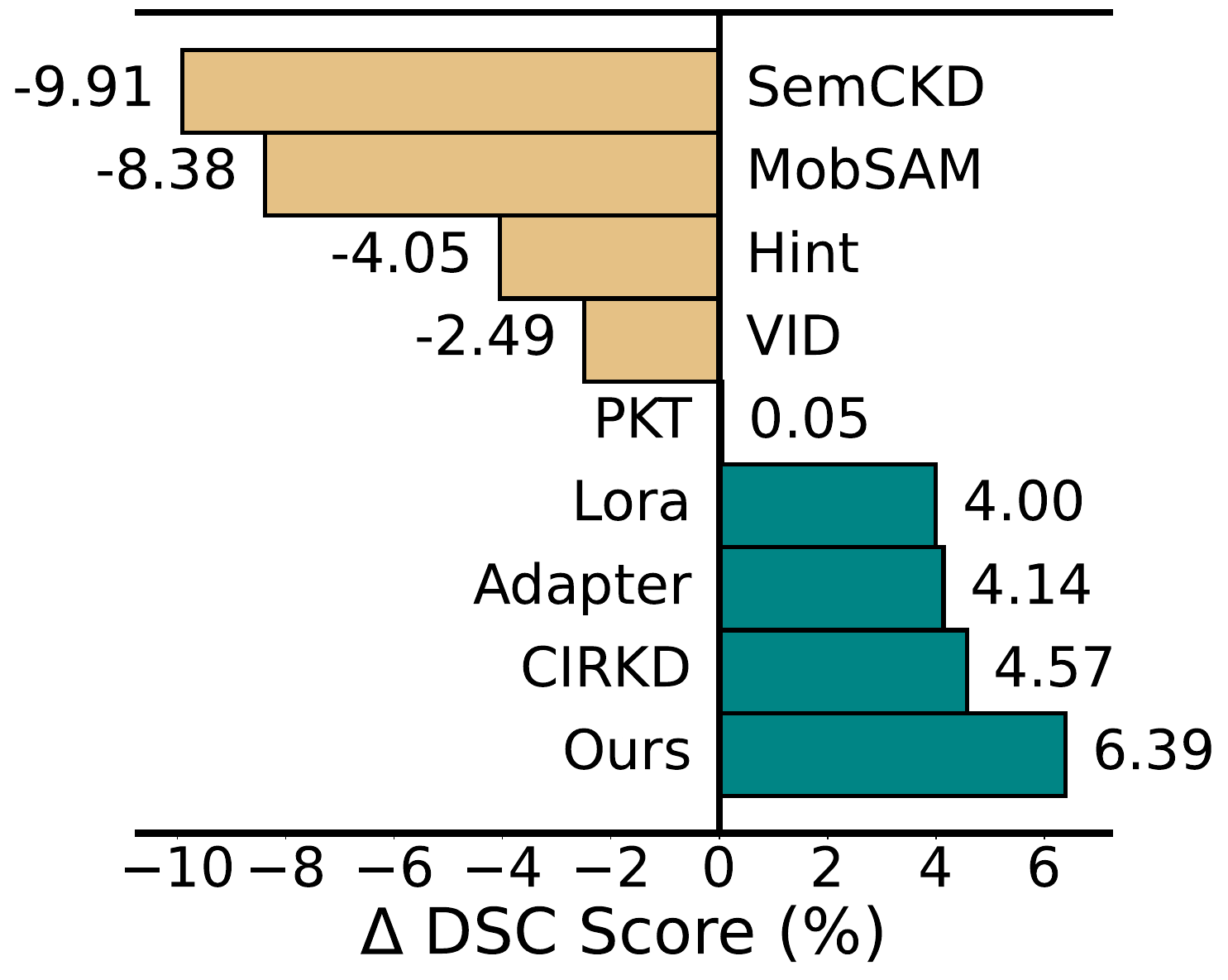}}
    \end{minipage}
    \hfill % 添加水平间距
    \begin{minipage}{0.24\textwidth}
        {\includegraphics[width=\textwidth]{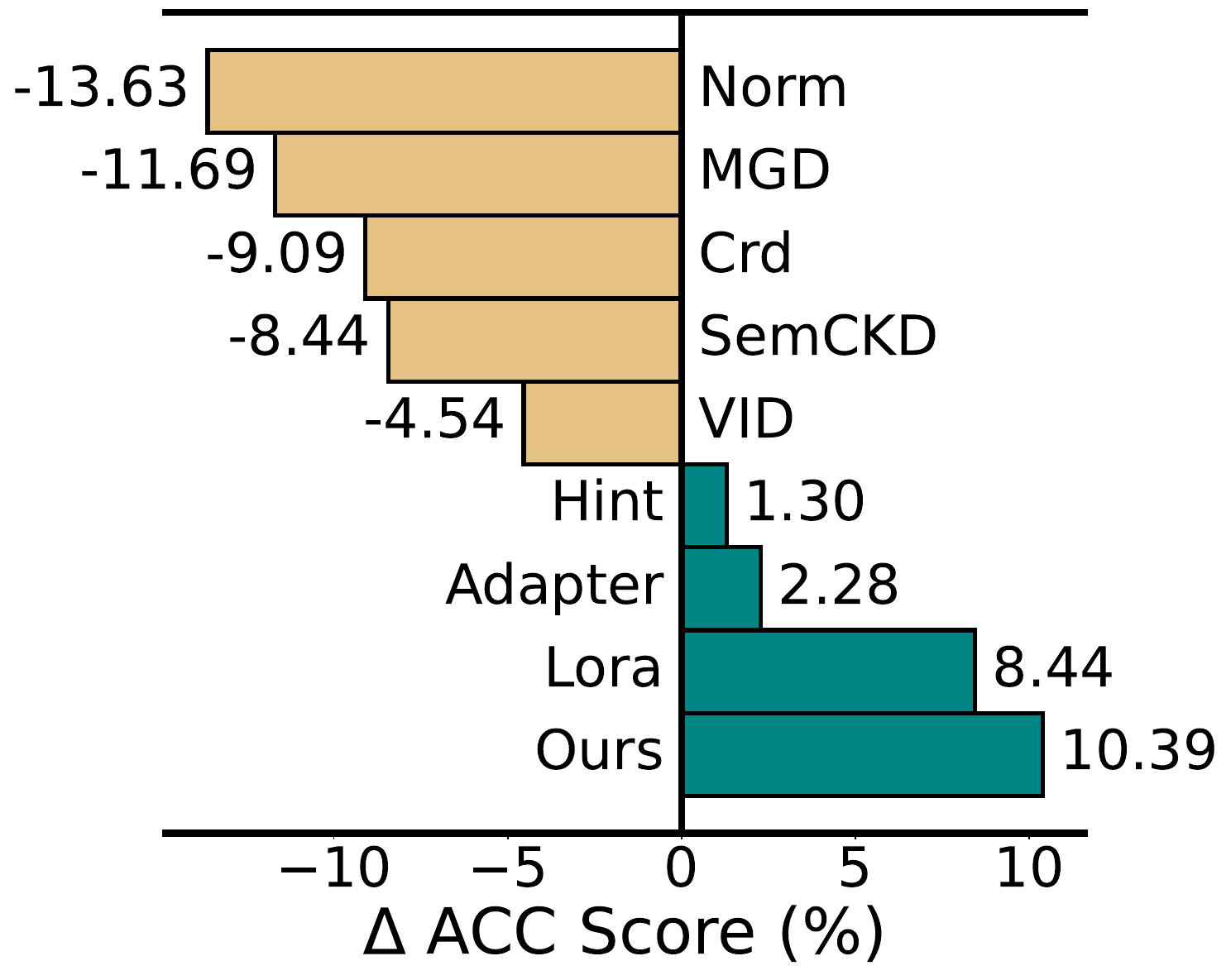}}
    \end{minipage}
    \caption{Comparison of methods in handling inconsistencies. The central axes in both figures represent the performance of the vanilla downstream model. The left figure shows performance gap when transferring knowledge from MedSAM (ViT-based) to ResNet18 (CNN-based) under structural inconsistencies. The right figure shows the gap when transferring from LVM-Med (self-supervised) to ResNet18 (supervised) under training strategy inconsistencies.}
    \label{future_study}
\end{figure}

\begin{figure}[t]
    \centering
    \includegraphics[width=0.92\linewidth]{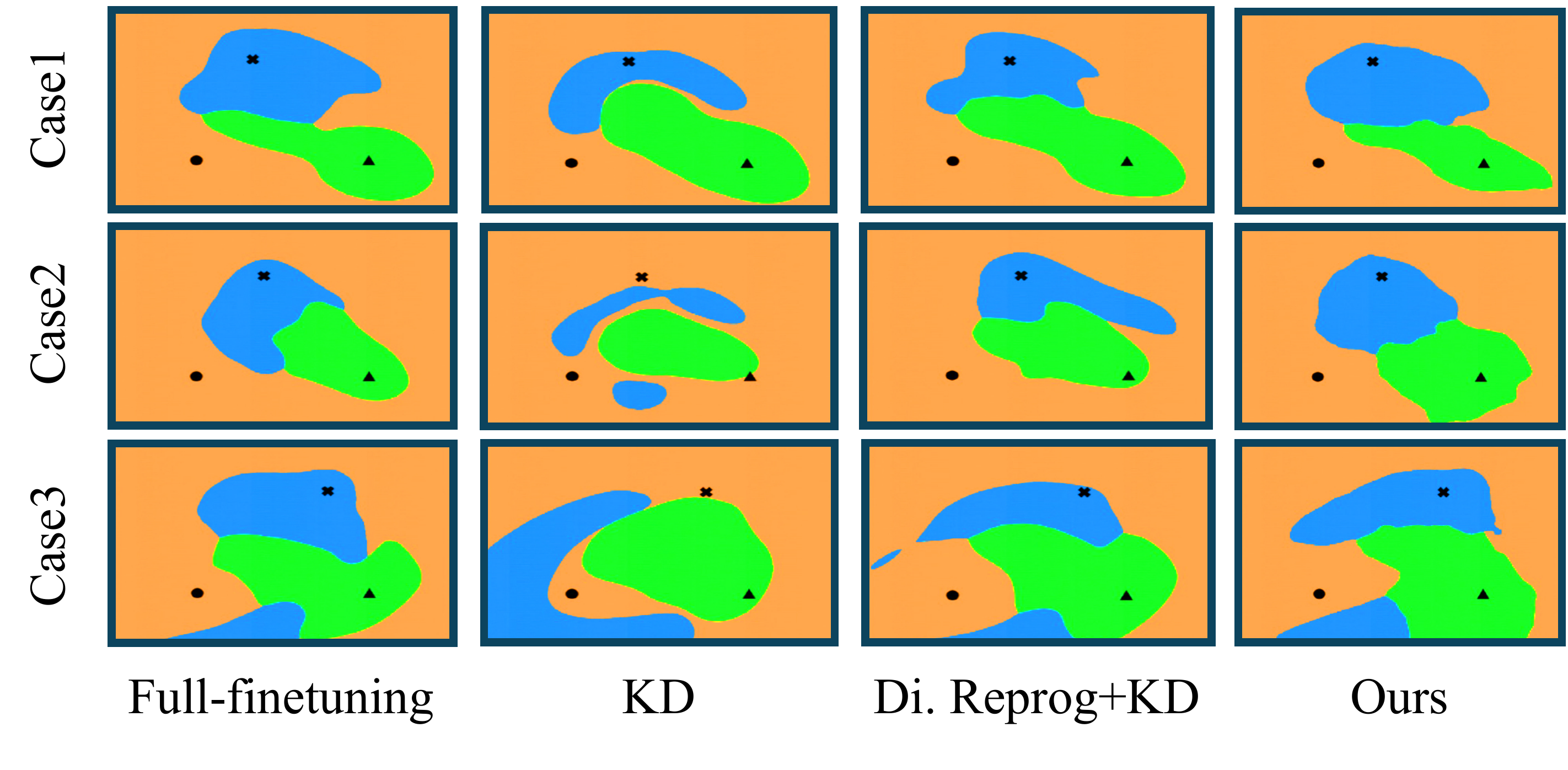}
    \caption{Comparison of decision boundaries of different methods on BUSI. In these plots, each color represents a class.}
    \label{fig:decision}
\end{figure}
\subsubsection{Mitigating Inconsistencies} 
Similar to the observation in section~\ref{sec:moti}, we further compare different methods in addressing inconsistencies between foundation and downstream models. 
As shown in Fig.~\ref{future_study}, in the presence of structural or training strategic inconsistencies, previous methods often fail to significantly improve the performance of the student model, with some methods even having detrimental effects. 
This reveals that the knowledge transferred by previous methods that ignore inconsistencies offer limited performance gains and sometimes produce adverse effects.
In contrast, DRD effectively mitigates these inconsistencies by reprogramming the knowledge from medical foundation model tailored to downstream task, promoting a beneficial knowledge transfer.

% \begin{figure}[t!]
% \centerline{\includegraphics[width=\linewidth]{img/Figure_1.png}}
% \caption{Comparing experiments on the effect of depth in classification and segmentation cases. Solid lines: Classification case (Accuracy); Dashed lines: Segmentation case (Dice coefficient)}
% \label{fig:ablation1}
% \end{figure}

% \subsection{Real-World Deployment and Evaluation}
% To further validate the practical applicability of our method, we deployed it across multiple 3D CT image diagnostic classification scenarios at Sir Run Run Shaw Hospital. 
% %
% Considering the constraints of clinical environments, our solution was designed to achieve rapid inference and high diagnostic accuracy despite limited computational resources. Initially, we selected a 3D diagnostic model pretrained on a large-scale public CT dataset as the foundational model. Leveraging our DRD method, we efficiently adapted this model to a variety of real-world diagnostic tasks, resulting in a series of lightweight diagnostic models. These optimized models, tailored for fast inference and ease of deployment, were subsequently implemented in various clinical departments to address specific diagnostic requirements.
%需要实际的，部署实验？可视化展示？在降低了多少计算成本的同时，达到了多少的准确率。

\subsubsection{Decision Boundaries Comparison in Classification Models} 
% Inspired by~\cite{somepalli2022can}, we also visualize the decision boundaries of classification models under different adaptation methods in Fig.~\ref{fig:decision}.

Inspired by the visualization method in~\cite{somepalli2022can}, where decision boundaries between classes are visualized by the borders between the colored regions, we also visualize the decision boundaries of classification models under different adaptation methods in Fig.~\ref{fig:decision}.
It can be seen that DRD has decision boundaries similar to those of fully fine-tuned foundation models compared with other baselines, even though in DRD, the vanilla foundation model backbone is not tuned. It proves that DRD can smartly transform the prior in medical foundation model into highly relevant knowledge for downstream tasks and transfer it efficiently, achieving similar effect of fully fine-tuning but in a lightweight manner.

\subsubsection{Segmentation Results Comparison in Segmentation Models} 

We visualize the segmentation results of DRD and other methods for both nuclear and organ segmentation tasks, comparing them with the ground truth, as shown in Fig.~\ref{fig：case_study}. As observed in the first row of images, all methods roughly produce segmentation masks similar to the ground truth. However, in the second row, where the segmentation results are magnified to reveal finer details, we find that, compared to previous methods, DRD demonstrates a clear advantage by producing segmentation masks that are more accurate and better aligned with the ground truth.

\section{Conclusion}
To construct efficient and computation-friendly adaptation of medical foundation models in the downstream tasks, it is crucial to address knowledge transfer challenges arising from inconsistencies in data distribution, model structure and practical constraints.
Different from previous PEFT methods and KD methods that cannot simultaneously address the domain, task and structure inconsistency and achieve lightweight deployment according to different real-world demands, we propose a novel DRD framework designed to overcome this challenge. By conducting a large number of experiments across various medical application scenarios, including 2D/3D classification and 2D/3D segmentation tasks, we demonstrate the generalizability and effectiveness of our method. In the future, we will explore the in-hospital applications by means of DRD with the state-of-the-art medical foundation models.

\appendices
\rev{
\section{Convergence Analysis under Different Reprogramming Depths}
\label{app:conv_n}

We further examine the optimization behavior of DRD when the reprogramming depth $N$ increases. Fig.~\ref{fig:n_convergence_appendix} shows the training total loss and validation Dice curves under $N=1,2,3,4$ on USC with MedSAM as the teacher and ViT-Tiny as the student. All tested settings exhibit steadily decreasing training loss and improving validation Dice in the early stage, followed by stable saturation rather than divergence. This suggests that increasing the reprogramming depth does not introduce noticeable convergence instability, and the default $N=4$ setting maintains smooth optimization while achieving favorable late-stage validation performance.

\begin{figure}[h]
    \centering
    \includegraphics[width=\columnwidth]{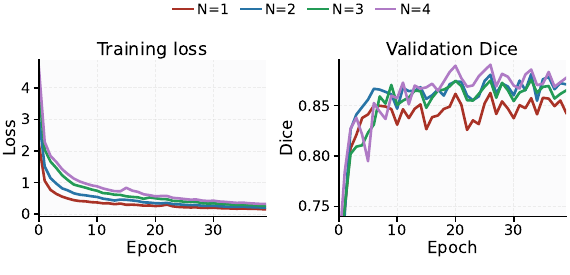}
    \caption{\rev{Convergence curves of DRD under different reprogramming depths $N$. Left: training total loss; right: validation Dice.}}
    \label{fig:n_convergence_appendix}
\end{figure}

\section{Parameter-level Gradient Diagnosis}
\label{app:grad_diag}

We also diagnose whether using a larger reprogramming depth introduces conflicting gradients on the late student layers. The analysis is conducted on USC with MedSAM as the teacher and four representative students. For each minibatch, we separately compute the gradients induced by $\mathcal{L}_{hybrid}$, $\mathcal{L}_{KD}$, and $\mathcal{L}_{CKA}$ on the last shared student encoder parameters, and compare them with the supervised gradient from $\mathcal{L}_{sup}$.
As shown in Table~\ref{tab:grad_diag_appendix}, the gradients from $\mathcal{L}_{hybrid}$ and $\mathcal{L}_{KD}$ have positive cosine similarity with the supervised gradient across all four students, indicating that they are cooperative rather than destructively opposed to the task objective. The CKA gradient on the last shared block is also extremely small relative to the supervised gradient (0.00\%--0.39\%), which is consistent with its role as an intermediate feature-alignment regularizer. Therefore, even when a larger reprogramming depth is used, the late student layers are not optimized by repeated conflicting updates; instead, they receive a single total gradient whose main auxiliary components remain task-aligned.
\begin{table}[t]
    \centering
    \caption{\rev{Gradient diagnosis on the last shared student encoder parameters (USC, teacher: MedSAM, default $N=4$). ``CKA/Sup'' denotes the ratio between the gradient norm of $\mathcal{L}_{CKA}$ and that of $\mathcal{L}_{sup}$ on the last shared student parameters.}}
    \label{tab:grad_diag_appendix}
    \setlength{\tabcolsep}{4pt}
    \renewcommand{\arraystretch}{1.05}
    \resizebox{0.90\columnwidth}{!}{%
    {\rev{
    \begin{tabular}{lccc}
        \toprule
        \textbf{Student} & \makecell{\textbf{Hybrid vs. Sup} \\ \textbf{cosine}} & \makecell{\textbf{KD vs. Sup} \\ \textbf{cosine}} & \makecell{\textbf{CKA/Sup} \\ \textbf{on last block}} \\
        \midrule
        ResNet18   & 0.541 & 0.552 & 0.39\% \\
        ShuffleNet & 0.399 & 0.397 & 0.00\% \\
        MobileNet  & 0.406 & 0.309 & 0.14\% \\
        ViT-Tiny   & 0.454 & 0.442 & 0.28\% \\
        \bottomrule
    \end{tabular}}}}
\end{table}

}

{
\bibliographystyle{IEEEtran}
\bibliography{reference}
}

\section*{Biography Section}
\vspace{-2cm}
\begin{IEEEbiography}
[{\includegraphics[width=1in,height=1.25in,clip,keepaspectratio]{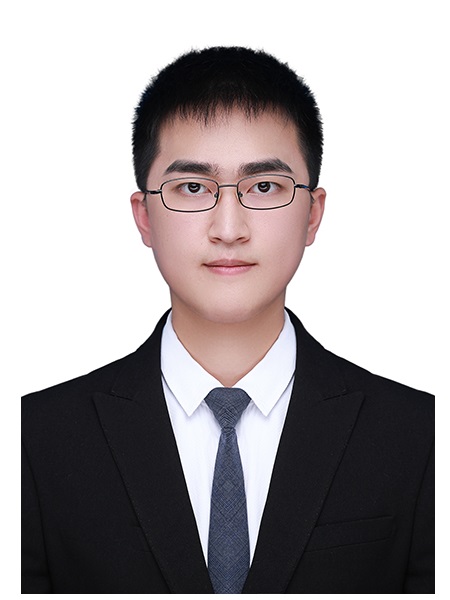}}]{Siyuan Du} 
    received a B.S. degree from University of Electronic Science and Technology of China, in 2023. 
    He is currently working toward the PhD degree from Fudan University, advised by Prof. J. Yao and Prof. Y. Zhang.  
    His research interests include computer vision and AI for healthcare.
\end{IEEEbiography}
\vskip -2\baselineskip plus -1fil
\begin{IEEEbiography}[{\includegraphics[width=1in,height=1.25in,clip,keepaspectratio]{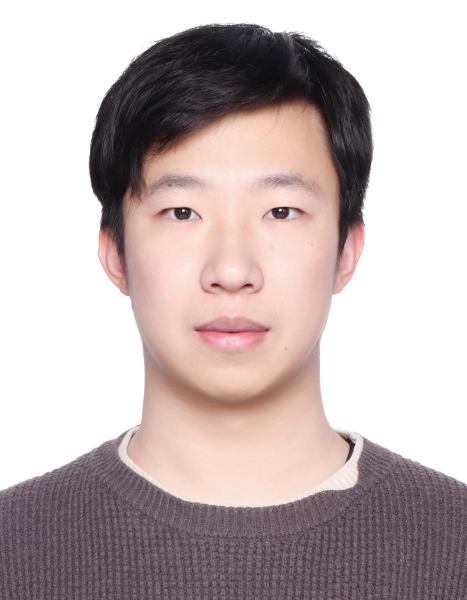}}]{Yuhang Zhou}
	received a B.S. degree from University of Electronic Science and Technology of China, in 2019. 
    He is currently working toward the PhD degree from Shanghai Jiao Tong University, advised by Prof. J. Yao and Prof. Y. Zhang.  
    His research interests include computer vision, machine learning and AI for healthcare.
\end{IEEEbiography}
\vskip -2\baselineskip plus -1fil
\begin{IEEEbiography}
[{\includegraphics[width=1in,height=1.25in,clip,keepaspectratio]{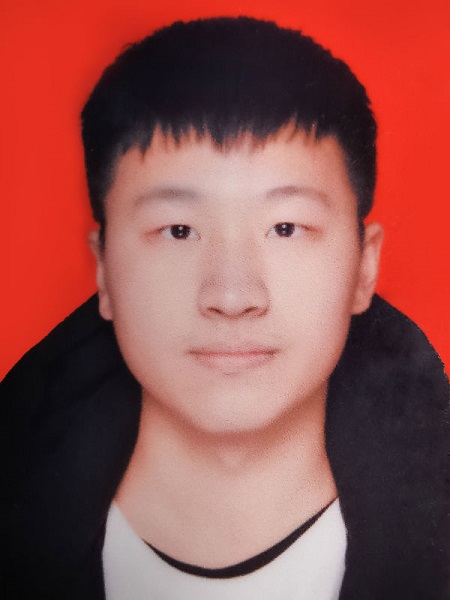}}]{Haolin Li}
	received a B.S. degree from University of Electronic Science and Technology of China, in 2023. 
    He is currently working toward the PhD degree from Fudan University, advised by Prof. J. Yao and Prof. Y. Zhang.  
    His research interests include computer vision and AI for healthcare.
\end{IEEEbiography}

\vskip -2\baselineskip plus -1fil
\begin{IEEEbiography}[{\includegraphics[width=1in,height=1.25in,clip,keepaspectratio]{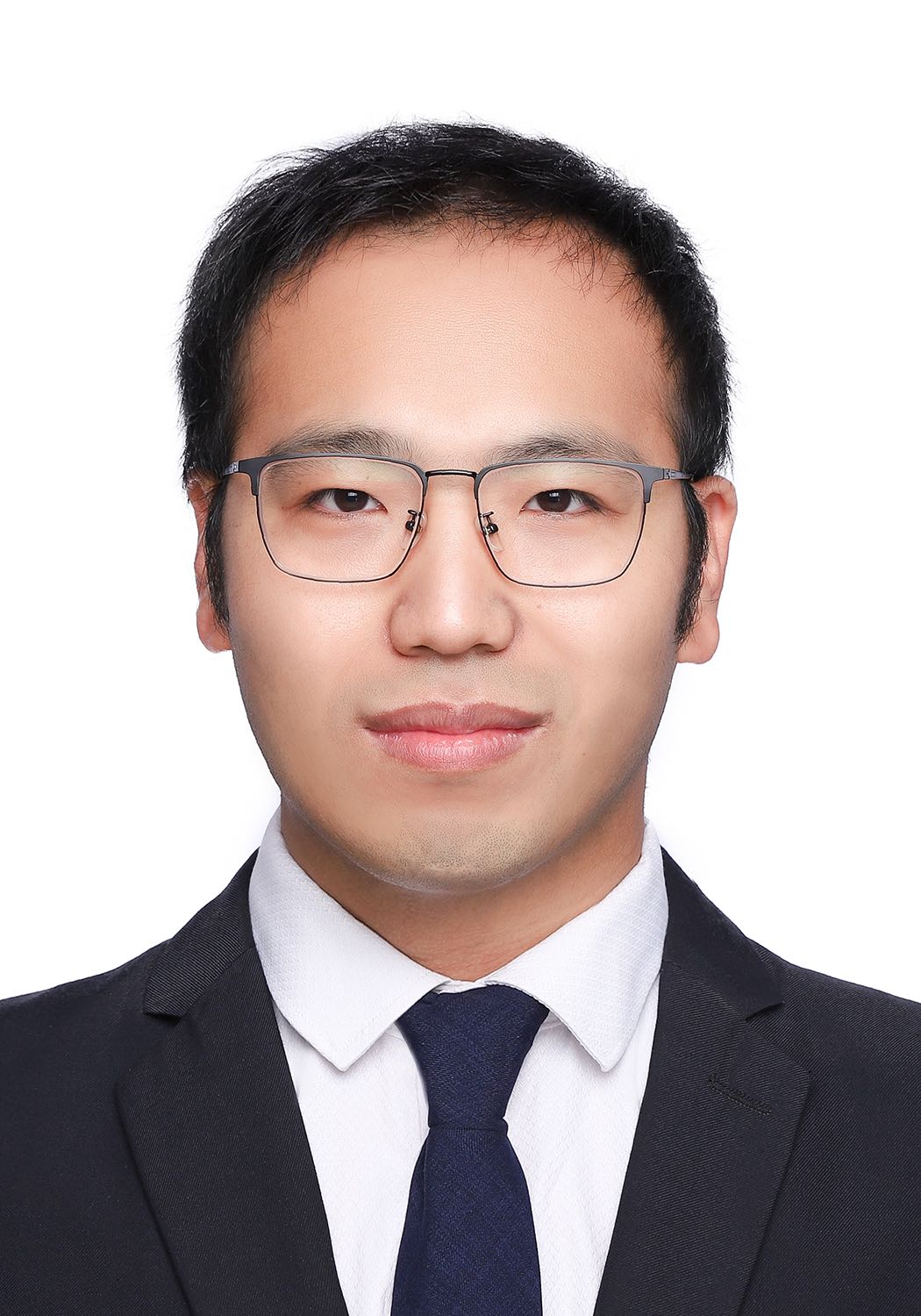}}]{Jiangchao Yao} is an Assistant Professor of Shanghai Jiao Tong University, Shnaghai China. 
He received the B.S. degree in information engineering from South China University of Technology, Guangzhou, China, in 2013. He got a dual Ph.D. degree under the supervision of Ya Zhang in Shanghai Jiao Tong University and Ivor W. Tsang in University of Technology Sydney.
His research interests include deep representation learning and robust machine learning.
\end{IEEEbiography}

\vskip -2\baselineskip plus -1fil
\begin{IEEEbiography}[{\includegraphics[width=1in,height=1.25in,clip,keepaspectratio]{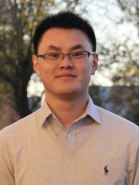}}]{Haishuai Wang} is a Ph.D. supervisor at the School of Computer Science and Technology, Zhejiang University, and a dual-appointed professor at the Second Affiliated Hospital of Zhejiang University School of Medicine. He received his Ph.D. from the Australian Artificial Intelligence Institute and the University of Technology Sydney, with a joint Ph.D. program at Washington University in St. Louis. He was also a postdoctoral researcher at Harvard University. Previously, he was an Assistant Professor at Fairfield University and a Research Assistant Professor at Harvard Medical School. His main research interests include knowledge mining and discovery, AI for healthcare, and large-scale medical models.

\end{IEEEbiography}

\vskip -2\baselineskip plus -1fil
\begin{IEEEbiography}[{\includegraphics[width=1in,height=1.25in,clip,keepaspectratio]{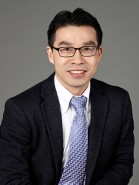}}]{Hui Lin} received his B.S., M.S., and Ph.D. degrees from Zhejiang University, Zhejiang, China. He is currently the Associate Dean of the College of Biomedical Engineering and Instrument Science, Zhejiang University, and a Chief Physician in General Surgery at the Sir Run Run Shaw Hospital, Zhejiang University.  His research interests mainly include the pathogenesis and diagnostic/treatment technologies of hepatobiliary and pancreatic tumors, surgical navigation, tumor-targeted therapy, and smart healthcare.
\end{IEEEbiography}

\vskip -2\baselineskip plus -1fil
\begin{IEEEbiography}[{\includegraphics[width=1in,height=1.25in,clip,keepaspectratio]{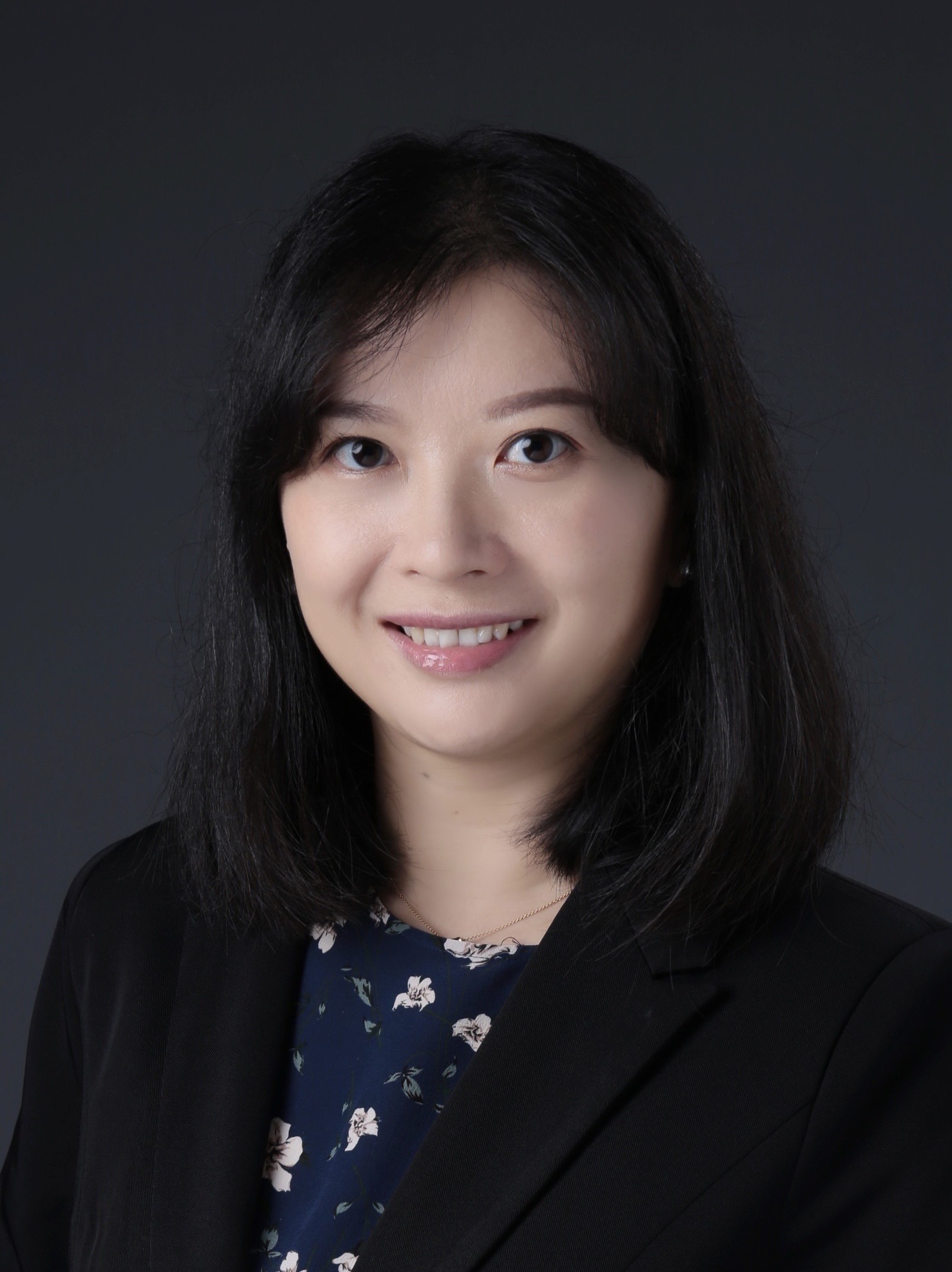}}]{Ya Zhang} (Member, IEEE) received the B.S. degree from Tsinghua University and the Ph.D. degree in information sciences and technology from the Pennsylvania State University. Since March 2010, she has been a professor with Cooperative Medianet Innovation Center, Shanghai Jiao Tong University. Prior to that, she worked with Lawrence Berkeley National Laboratory, University of Kansas, and Yahoo! Labs. Her research interest is mainly on data mining and machine learning, with applications to information retrieval, web mining, and multimedia analysis.
\end{IEEEbiography}
\vskip -2\baselineskip plus -1fil
\begin{IEEEbiography}[{\includegraphics[width=1in,height=1.25in,clip,keepaspectratio]{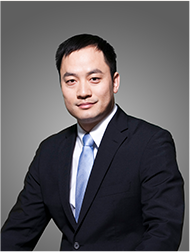}}]{Yanfeng Wang} received the B.E. degree in information engineering from the University of PLA, Beijing, China, and the M.S. and Ph.D. degrees in business management from the Antai College of Economics and Management, Shanghai Jiao Tong University, Shanghai, China. He is currently the Vice Director of the Cooperative Medianet Innovation Center and also the Vice Dean of the School of Electrical and Information Engineering, Shanghai Jiao Tong University. His research interests mainly include media big data and emerging commercial applications of information technology.
\end{IEEEbiography}

\end{document}